\documentclass[12pt,english]{article}
\usepackage{tikz}
\usetikzlibrary{arrows.meta}
\usepackage{booktabs} 
\usepackage{mathptmx}
\usepackage{geometry}
\usepackage{xcolor}
\usepackage{array}
\usepackage{bbding}
\usepackage{multirow}
\usepackage[fleqn]{amsmath}
\usepackage{amssymb}
\usepackage{amsthm}
\usepackage{graphicx}
\usepackage{natbib}
\usepackage{lscape}
\usepackage[hyphens]{url}
\usepackage[hidelinks]{hyperref}
\usepackage{comment}
\usepackage{bm}
\usepackage[title]{appendix}
\usepackage{longtable}
\usepackage{mathtools}
\usepackage{chngcntr}
\usepackage{authblk}
\usepackage{babel}
\usepackage{dsfont}
\usepackage{subcaption}
\usepackage{fancyhdr}
\usepackage{cleveref}
\usepackage{abstract}

\makeatletter
\allowdisplaybreaks
\date{}

\fancypagestyle{plain}{%
	\fancyhf{} 
	
	\chead{\footnotesize \textcolor{gray}{\textit{\shortauthors -- \runningtitle}}}
	\rhead{\footnotesize \textcolor{gray}{\thepage}}
}
\makeatletter
\def\blfootnote{\xdef\@thefnmark{}\@footnotetext}
\makeatother

\makeatletter
\def\titlepageext{
	\begin{center}	
	{\parindent0pt
		\rule{0.9\textwidth}{1pt}
		\begin{minipage}[t]{0.25\textwidth}
			\small {\it Keywords:}\\
			\keyword
		\end{minipage}%
		\hspace{3mm}
		\begin{minipage}[t]{0.6\textwidth}
			\small \abstract
		\end{minipage}%
		
		\rule{0.9\textwidth}{2pt}
	}
	\end{center}

	\blfootnote{* Corresponding author. E-mail address: \href{mailto:\corresemail}{\corresemail}.}
}
\makeatother

\usepackage[mathlines, pagewise]{lineno}
\usepackage{etoolbox} 

\newcommand*\linenomathpatchAMS[1]{%
	\expandafter\pretocmd\csname #1\endcsname {\linenomathAMS}{}{}%
	\expandafter\pretocmd\csname #1*\endcsname{\linenomathAMS}{}{}%
	\expandafter\apptocmd\csname end#1\endcsname {\endlinenomath}{}{}%
	\expandafter\apptocmd\csname end#1*\endcsname{\endlinenomath}{}{}%
}

\expandafter\ifx\linenomath\linenomathWithnumbers
\let\linenomathAMS\linenomathWithnumbers
\patchcmd\linenomathAMS{\advance\postdisplaypenalty\linenopenalty}{}{}{}
\else
\let\linenomathAMS\linenomathNonumbers
\fi

\linenomathpatchAMS{gather}
\linenomathpatchAMS{multline}
\linenomathpatchAMS{align}
\linenomathpatchAMS{alignat}
\linenomathpatchAMS{flalign}

\nolinenumbers

\usepackage{mathrsfs}
\usepackage{amssymb}
\usepackage{amsmath}
\usepackage{booktabs}
\usepackage{tabularx}
\usepackage{multirow}
\usepackage{makecell}
\usepackage{graphicx}
\usepackage{subcaption}

\newcolumntype{L}{>{\raggedright\arraybackslash}X}
\newcolumntype{M}[1]{>{\centering\arraybackslash}m{#1}}

\title{Machine Unlearning of Traffic State Estimation and Prediction}

\def\shortauthors{Wang et al.}
\def\runningtitle{Machine Unlearning of TSEP}

\author[a]{Xin Wang}
\author[b]{R. Tyrrell Rockafellar}
\author[a$\ast$]{Xuegang(Jeff) Ban}
\affil[a]{Department of Civil and Environmental Engineering, University of Washington, Seattle, WA, 98195, United States}
\affil[b]{Department of Mathematics, University of Washington, Seattle, WA, 98195, United States}
\def\corresemail{banx@uw.edu}

\def\abstract{Data-driven traffic state estimation and prediction (TSEP) relies heavily on data sources that contain sensitive information. While the abundance of data has fueled significant breakthroughs, particularly in machine learning-based methods, it also raises concerns regarding privacy, cybersecurity, and data freshness. These issues can erode public trust in intelligent transportation systems. 
Recently, regulations have introduced the "right to be forgotten", allowing users to request the removal of their private data from models. As machine learning models can remember old data, simply removing it from back-end databases is insufficient in such systems.

To address these challenges, this study introduces a novel learning paradigm for TSEP—\textit{Machine Unlearning TSEP}—which enables a trained TSEP model to selectively forget privacy-sensitive, poisoned, or outdated data. By empowering models to "unlearn," we aim to enhance the trustworthiness and reliability of data-driven traffic TSEP.}

\def\keyword{TSEP\\Privacy\\Machine Unlearning}

\begin{document}
\maketitle
\titlepageext

\section{Introduction}
Traffic State Estimation and Prediction (TSEP) has been extensively studied to reconstruct traffic state variables (e.g., flow, density, speed, travel time, etc.) using (partial) observed traffic data~\citep{antoniou2013dynamic,ban2011real,shi2021physics,li2020real}.  In recent years, advancements in data collection technologies have enabled TSEP methods to integrate traffic data from diverse sources for more accurate and robust estimation and prediction~\citep{wang2016efficient,makridis2023adaptive}. These data sources can be broadly categorized into infrastructure-collected data and user-contributed data. Infrastructure-collected data typically includes information collected from loop detectors, traffic cameras, and radars installed on roadways or at intersections. In contrast, user-contributed data is derived from individuals, often through vehicles or personal devices, such as GPS traces, vehicle trajectories, and probe data collected via mobile apps or in-vehicle systems. Additionally, V2X communication data~\citep{hasan2020securing}, which bridges these two categories, involves information exchanged among vehicles, other road users, and infrastructure, or directly between vehicles. 

Although the integration of diverse data has significantly improved the accuracy and reliability of TSEP, it also raises concerns about legal, privacy, and safety violations stemming from adversarial data manipulation (e.g., data poisoning attacks), datasets containing privacy sensitive information, and incorrect or outdated training data~\citep{wang2024data,de2013unique}. For example, GPS traces or trajectory data linked to specific vehicles could accidentally expose sensitive locations of individuals, posing significant privacy and security risks to the vehicle owners~\citep{iqbal2010privacy}. In addition, outdated or poisoned training data can lead to inaccurate predictions~\citep{wang2024data,wang2024transferability}, which undermines the reliability and trustworthiness of machine learning models. As machine learning (ML) and deep learning (DL) models are known to potentially memorize training data, only removing data from back-end databases (without retraining the model) is often not sufficient to remove the impacts of the problematic data. A straightforward solution to these issues might involve simply removing the sensitive or poisoned data and retraining the model from scratch. However, as ML and DL models grow increasingly large and complex, retraining can be prohibitively expensive, requiring vast computational resources and significant time. Thus, it is impractical to retrain models every time data needs to be removed, underscoring the demand for efficient learning approaches that comply with data deletion requests. This has become more pressing recently due to regulations such as the European Union’s General Data Protection Regulation (GDPR), the California Consumer Privacy Act (CCPA), and Canada's Personal Information Protection and Electronic Documents Act (PIPEDA). These regulations enforce the “right to be forgotten”~\citep{regulation2016regulation}, which requires organizations to delete personal data on request and to take reasonable steps to erase the impacts of such data from their systems, including any ML/DL models trained using the data. 

The process of deleting data and its impacts from an ML or DL model, without retraining, is known as \textit{machine unlearning}~\citep{bourtoule2021machine, nguyen2022survey,triantafillou2024we}. 
It aims to produce a modified model, referred to as the \textit{unlearned model}, that is equivalent to, or at least behaves like, a model retrained on the remaining data after the removal request has been processed (which we refer to as the \textit{gold standard model}). In this study, we focus on the machine unlearning problem in data-driven TSEP methods. Such methods leverage ML and DL techniques, such as neural networks, to uncover dependencies in historical traffic data~\citep{antoniou2013dynamic,li2017diffusion}. Examples include Least Squares (LS) models to estimate delay patterns using mobile data~\citep{ban2011real}, Support Vector Machines (SVMs) applied to vehicle classification using vehicle trajectory data~\citep{sun2013vehicle}, and long short term memory (LSTM) neural network for traffic flow prediction~\citep{li2020real}. Recently, physics-informed neural networks (PINNs) have emerged as a novel research trend for TSEP that combines traffic flow theory with learning-based approaches~\citep{shi2021physics, lu2023physics,thodi2022incorporating}. This integration seeks to overcome the "black-box" nature of neural networks applied by data-driven methods, offering more interpretable solutions that align closely with fundamental traffic principles and domain knowledge. Note that both classical ML-based and PINN-based TSEP methods often incorporate problem-specific constraints. For instance, a model estimating vehicle queue lengths at intersections must consider the upper and lower limits of the signal's green time and the physical car-following behavior between vehicles~\citep{ban2011real}. Additionally, PINN-based TSEP methods must ensure that their solutions adhere to traffic flow theories described by physical models, such as the Lighthill-Whitham-Richards (LWR) model~\citep{shi2021physics}.

After receiving a data deletion request, machine unlearning can "erase" the influence of the traffic data to be forgotten (e.g., a subset of trajectory data) from the trained TSEP model, adjusting the model’s solution (e.g., neural network weights) towards the solution of the gold standard model. Existing machine unlearning frameworks often assumes nonexistence of contraints and implement the removal by directly estimating the solution change caused by the data removal. Further discussion is provided later in Section \ref{sec:review_MUL}. However, when the problem involves constraints (e.g., traffic flow conservation laws or physical car-following behavior), the solution space is restricted, and the unlearning process must ensure that the adjusted solution remains feasible under these constraints. Furthermore, since the constraints are often data-dependent, data removal can also modify the feasible region. This requires measuring the influence of the removed data on both the objective function and the constraints simultaneously to ensure that the solution of unlearned model is optimal and valid within the updated feasible region. To the best of our knowledge, there is no work that has explored machine unlearning in the context of constrained problems. 

Assume our TSEP models rely on the training dataset $D=\left[z_1, z_2, \ldots, z_N\right]$ to determine the optimal solution set $\theta$. Consider a request to remove data points $z_1,z_2$. We assign a weighting vector $\eta=[\eta_1,\eta_2,\dots,\eta_N]$ to the data points, and define a solution mapping $S(\eta;D)=\{\theta \mid$ $\theta$ satisfies the TSEP model $\}$. By gradually reducing $\eta_{1}$ and $\eta_{2}$ from 1 to 0, we can diminish the influence of $z_{1}$ and $z_{2}$ on the solution set. The key to avoiding retraining lies in measuring how the solution $\theta$ evolves as $\eta$ changes.  We address this challenge through a sensitivity analysis of the TSEP by defining proper concepts, measures, and tools that can properly deal with constrainted learning problems. Sections~\ref {sec:2} and \ref{sec:3} will provide more discussions on this. Our main contributions are as follows: 
 
1) We propose a sensitivity-analysis-based machine unlearning framework for constrained learning models, with unconstrained models included as a special case.

2) We tailor the proposed unlearning method to various TSEP problems and provide a solution algorithm based on quadratic programming.

3) Experiments demonstrate that the unlearned model matches the performance of retraining, but at a fraction of the computational cost.

The remainder of the paper is organized as follows.
Section~\ref{sec:related work} reviews related work on TSEP and machine unlearning approaches.
Section~\ref{sec:2} introduces the theoretical foundation of our machine unlearning method by formulating it as a sensitivity analysis problem for constrained optimization.
Section~\ref{sec:3} customizes the unlearning method for TSEP problems and proposes a numerical algorithm based on quadratic programming.
Section~\ref{sec:4} evaluates the proposed algorithm on two tasks: SVM-based vehicle classification and PINN-based velocity field reconstruction.
Section~\ref{sec:5} concludes the paper and discusses directions for future work.

\section{Related Work} \label{sec:related work}
\subsection{TSEP}
Traffic State Estimation and Prediction (TSEP) aims to reconstruct and forecast key traffic variables (e.g., flow, speed, and density) using historical and real-time data. Existing methods fall broadly into two categories: \textit{model-based approaches} and \textit{data-driven approaches}. The model-based approaches utilize physics-based models, such as the traffic flow and car-following models, to estimate traffic states. Examples include 
Kalman Filtering~\citep{wang2005real} and Cell Transmission Model~\citep{daganzo1995cell}. This paper focuses on the data-driven approaches, which learn traffic variables directly from data. Representative data-driven methods include convolutional neural networks (CNNs)~\citep{bogaerts2020graph}, recurrent neural networks (RNNs)~\citep{li2020real}, and graph-based models such as DCRNN~\citep{li2017diffusion} and STGCN~\citep{han2020stgcn}. While physical models provide transparent structures grounded in traffic flow theory, they often struggle to capture the complexity of real-world conditions, especially when applied to unobserved scenarios. In contrast, data-driven models (e.g., deep neural networks) can learn rich representations from historical data and achieve superior predictive accuracy, but they typically lack interpretability and may violate known physical laws, limiting their trustworthiness in safety-critical applications. Recently, PINNs~\citep{shi2021physics,lu2023physics,raissi2017physics} have emerged as a promising hybrid framework that embeds physical models into deep neural networks. The learned solution is data-driven while remaining aligned with physical principles, thereby enhancing both interpretability and robustness. For example, \citet{shi2021physics} reconstructs the velocity field using a multilayer perceptron, while ensuring that the estimated velocity follows the Greenshields-based LWR model. The proposed model incorporates a regularization term based on the residual error of the LWR partial differential equation (PDE). \citet{tang2024physics} leverages an encoder–decoder architecture to calibrate traffic flow model parameters using speed and flow observations. The decoder approximates the physical model, enabling the model to generate reasonable traffic flow parameters. Given that TSEP is broad and rapidly evolving, this paper does not attempt to provide a comprehensive review, but instead focuses on the machine unlearning within the TSEP context.

\subsection{Machine Unlearning}\label{sec:review_MUL}
Since the 2020s, machine unlearning has emerged as a promising solution to address growing concerns over privacy, security, robustness, and bias reduction. It can correct the model to "forget" privacy-related or corrupted data without requiring a full retraining.


Existing machine unlearning methods can be broadly categorized into data-based and model-based approaches. Data-based methods leverage data partitioning techniques to facilitate efficient retraining. A representative example is the SISA framework~\citep{bourtoule2021machine}, which partitions the dataset into multiple slices. Each slice is used to train an individual model, and the final prediction is obtained by aggregating the outputs of these models. When unlearning a data point, only the affected slice is retrained, significantly reducing computation. Model-based methods, on the other hand, aim to modify the model parameters directly in response to data deletion. For linear models, this can be efficiently achieved using rank-one updates of matrix inverses~\citep{birattari1998lazy,cao2015towards}. For deep neural networks (DNNs), one common approach is based on influence functions~\citep{law1986robust,koh2017understanding}, which use Taylor expansions to approximate the impact of a removed data point on the model parameters. However, influence function based methods typically require explicit access to Hessian matrices and are limited to handling the removal of a single data point. More recently, \citet{golatkar2020forgetting} introduced a method inspired by differential privacy, which leverages the Fisher Information Matrix to effectively erase the influence of deleted training samples. Optimization-based approaches, such as DeltaGrad~\citep{wu2020deltagrad}, accelerate the retraining process by reusing and adjusting previously computed gradients. Since model retraining typically relies on gradient descent, computing accurate gradients is the most time-consuming step. They evaluate how small changes in the dataset (e.g., the removal of a few data points) influence the gradients, and leverage this information to approximate the updated gradients without recomputing them from scratch. 

To the best of our knowledge, existing machine unlearning methods primarily focus on unconstrained learning tasks, making them unsuitable for direct application to TSEP problems. This paper extends the concept of influence functions to constrained optimization settings and further allows for the removal of multiple data points simultaneously.

\section{Machine Unlearning as Sensitivity Analysis}\label{sec:2}
To study machine unlearning in the presence of constraints, we begin with a general constrained learning formulation:
\begin{equation}\label{eq:general}
\begin{aligned}
\bar{\theta} = \arg \min_{\theta} \quad & g_0(D, \theta) \\
\text{s.t.} \quad & g_j(D, \theta) \leq 0, \quad j = 1, \ldots, J, \\
& h_t(D, \theta) = 0, \quad t = 1, \ldots, T.
\end{aligned}
\end{equation}
Here, $D=[z_i]_{i=1}^N$ denotes a dataset, and \( g_0(\cdot) \) represents the objective function. The constraints \( g_i \) and \( h_j \) represent general inequality and equality constraints, respectively. In specific learning problems, they often represent application-specific requirements and constraints. For example, as shown in Section~\ref{sec:3}, ML-based TSEP problems may embed physical consistency, resource limits, or even requirements such as fairness. $\bar{\theta}$ denotes the optimal solution derived from the full dataset. 


Machine unlearning aims to remove a portion of user-sensitive information, denoted as $D^r$, from the full dataset $D$. We denote the solution derived from the reduced dataset $D\setminus D^r$ as $\bar{\theta}_{-D^r}$: 
\begin{equation}\label{eq:Gold standard model}
\begin{aligned}
\bar{\theta}_{-D^r}  = \arg \min_{\theta} \quad & g_0( D\setminus D^r, \theta) \\
\text{s.t.} \quad & g_j( D\setminus D^r, \theta) \leq 0, \quad j = 1, \ldots, J, \\
& h_t( D\setminus D^r, \theta) = 0, \quad t = 1, \ldots, T.
\end{aligned}
\end{equation}
We refer to the above problem as the \textit{gold standard model}, as it provides the ground truth solution after data removal. As aforementioned, the gold standard model implies retraining of the original model after data is removed, which may be computationally expensive. In the following sections, we demonstrate how to estimate $\bar{\theta}_{-D^r}$ using $\bar{\theta}$ without retraining. 

In the constrained learning~(\ref{eq:general}), each data point $z_i$ is assigned an equal weight. To remove the data points $ D^r$, we borrow the concept of the influence function from robust statistics~\citep{law1986robust}, which was originally designed for unconstrained problems. Specifically, we introduce a data weight vector $\eta=[\eta_1,\eta_2,\dots,\eta_N] \in \mathbb{R}^N$ into the problem~(\ref{eq:general}) , where $\eta_i$ denotes the weight of a data point $z_i$. The data weighted constrained learning problem is defined as: 
\begin{equation}\label{eq:data weighted tsep}
\begin{aligned}
\theta(\eta) = \arg \min_{\theta} \quad & g_0(\eta \cdot D, \theta) \\
\text{s.t.} \quad & g_j( \eta \cdot D, \theta) \leq 0, \quad j = 1, \ldots, J, \\
& h_t( \eta \cdot D, \theta) = 0, \quad t = 1, \ldots, T.
\end{aligned}
\end{equation}
The weight vector $\bar{\eta}=\vec{1}$ recovers the original problem~(\ref{eq:general}), i.e., $\bar{\theta} = {\theta}({\bar{\eta}})$. Removing the data $z_{i}$ is equivalent to setting its corresponding weight $\eta_{i}=0$. We denote the data weight corresponding to the gold standard model~(\ref{eq:Gold standard model}) as $\eta^r=\{\eta_i^r\}_1^N \in \mathbb{R}^N$, where
\begin{equation*}
\eta_i^r= \begin{cases}0, & \text { if } z_i \in D^r  \\ 1, & \text { otherwise }\end{cases}  \end{equation*}
It follows directly that $\bar{\theta}_{-D^r}=\theta(\eta^r)$. 

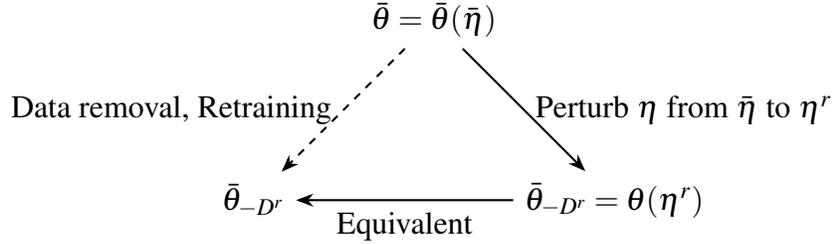
\begin{figure}[h!]
\centering
\begin{tikzpicture}[>=Stealth, thick, scale=1.2]

\node (top) at (0,2) {\( \bar{\theta} = \bar{\theta}({\bar{\eta}}) \)};
\node (left) at (-2,0) {\( \bar{\theta}_{-D^r} \)};
\node (right) at (2,0) {\( \bar{\theta}_{-D^r}=\theta(\eta^r) \)};

\draw[->] (top) -- (right) node[midway, right] {Perturb $\eta$ from $\bar{\eta}$ to $\eta^r$};
\draw[dashed,->] (top) -- (left) node[midway, left] {Data removal, Retraining};
\draw[<-, thick] (left) -- (right) node[midway, below] {Equivalent};

\end{tikzpicture}
\caption{Illustration of $\bar{\theta}_{-D^r}$ estimation without retraining} 
\label{fig: eta} 
\end{figure}

Introducing the weight vector transforms the machine unlearning problem into a sensitivity analysis with respect to the data weights $\eta$. By analyzing how the solution $\theta(\eta)$ changes when the weight vector shifts from the original full-data weight $\bar{\eta}=\vec{1}$ to the weight vector 
$\eta^r$, we can estimate solution $\bar{\theta}_{-D^r}=\theta(\eta^r)$ without retraining. The connection between sensitivity analysis and machine unlearning is depicted in Figure~\ref{fig: eta}.

\subsection{Sensitivity Analysis} \label{sec:2.1}
This section aims to analyze the sensitivity of the solution $\theta$ with respect to the data weights $\eta$. We begin by formulating the optimality condition of the data weighted learning problem~(\ref{eq:data weighted tsep}) as a variational inequality (VI). When the data weights change from $\bar{\eta}$ to $\eta^r$, the updated solution $\theta(\eta^r)$ must jointly satisfy the VI. This fact enables us to estimate the corresponding change in the solution given a change in the data weights. We further linearize this VI to reduce computational cost.

We first define the Lagrangian function of data weighted constrained learning~(\ref{eq:data weighted tsep}) as follows: 
\begin{equation}
\mathcal{L}(\eta, \theta, \lambda)=g_0(\eta \cdot D, \theta)+\sum_{j=1}^J \lambda_g^j g_j(\eta \cdot D, \theta)+\sum_{t=1}^T \lambda_h^t h_t(\eta \cdot D, \theta),
\end{equation}
where $\lambda=[\lambda_g^j,\lambda_h^t],\lambda_g^j \geq 0$  are the Lagrange multipliers associated with the constraints. The VI that captures the associated first-order optimality conditions is given by~\citep{luo1996mathematical,facchinei2003finite}:
\begin{equation}\label{VI}
f(\eta, \theta, \lambda) + N_E(\theta, \lambda) \ni 0,
\end{equation}
where
\begin{equation}
\left\{\begin{array}{l}
f(\eta, \theta, \lambda) = 
\left( 
\nabla_\theta \mathcal{L}(\eta, \theta, \lambda),\ 
- \nabla_\lambda \mathcal{L}(\eta, \theta, \lambda)
\right)^\top, \\
 E = \mathbb{R}^{dim(\theta)} \times \left[ \mathbb{R}_+^J \times \mathbb{R}^{T} \right].
\end{array}\right.
\end{equation}
Here $N_E(\theta, \lambda)$ denotes the normal cone to $E$ at $(\theta, \lambda)$; see its definition in~\cite{dontchev2009implicit} and also listed in Appendix~\ref{normal cone appendix}. $\bar{\theta}$ is the optimal solution associated with the full-data weight $\eta=\bar{\eta}$, and we denote its corresponding Lagrange multiplier as $\bar{\lambda}$. $(\bar{\eta},\bar{\theta},\bar{\lambda})$ satisfies the VI~(\ref{VI}), i.e., 
\begin{equation}
    f(\bar{\eta}, \bar{\theta}, \bar{\lambda}) + N_E(\bar{\theta}, \bar{\lambda}) \ni 0.
\end{equation}
When $\eta$ shifts from $\bar{\eta}$ to $\eta^r$, the change in the solution and Lagrangian multiplier, denoted by $\Delta \theta$ and $\Delta \lambda=[\Delta\lambda_g^j, \Delta\lambda_h^t]$, satisfy the following relation, due to \eqref{VI}:
\begin{equation}\label{VI2}
    f(\eta^r, \bar{\theta}+\Delta \theta, \bar{\lambda}+\Delta \lambda) + N_E(\bar{\theta}+\Delta \theta, \bar{\lambda}+\Delta \lambda) \ni 0.
\end{equation}
To reduce computational cost, we linearize the VI~(\ref{VI2}) as follows:
\begin{equation}\label{eq:linear VI}
f(\bar{\eta}, \bar{\theta}, \bar{\lambda})+\nabla_{\eta} f(\bar{\eta}, \bar{\theta}, \bar{\lambda})(\eta^r-\bar{\eta})+\nabla_{(\theta, \lambda)} f(\bar{\eta}, \bar{\theta}, \bar{\lambda})[\Delta \theta ; \Delta \lambda]+N_E(\bar{\theta}+\Delta \theta, \bar{\lambda}+\Delta \lambda) \ni \mathbf{0},
\end{equation}
where $\nabla_{(\theta, \lambda)} f(\bar{\eta}, \bar{\theta}, \bar{\lambda})$ is the Jocobian matrix:
\begin{equation}
\nabla_{(\theta, \lambda)} f(\bar{\eta}, \bar{\theta}, \bar{\lambda})=\left[\begin{array}{c}
\nabla_\theta^2 \mathcal{L}(\bar{\eta}, \bar{\theta}, \bar{\lambda}), \nabla_{\theta \lambda}\mathcal{L}(\bar{\eta}, \bar{\theta}, \bar{\lambda}) \\
-\nabla_{\theta \lambda} \mathcal{L}(\bar{\eta}, \bar{\theta}, \bar{\lambda}),-\nabla_{\lambda \lambda}\mathcal{L}(\bar{\eta}, \bar{\theta}, \bar{\lambda})
\end{array}\right].
\end{equation}
Following the theorem 2E.4 in~\cite{dontchev2009implicit}, the VI~(\ref{eq:linear VI}) can be simplified as: 
\begin{equation}\label{eq:auxiliary VI}
\nabla_\eta f(\bar{\eta}, \bar{\theta}, \bar{\lambda})(\eta^r-\bar{\eta})+\nabla_{(\theta, \lambda)} f(\bar{\eta}, \bar{\theta}, \bar{\lambda})[\Delta \theta ; \Delta \lambda]+N_{\bar{E}}(\Delta \theta, \Delta \lambda) \ni \mathbf{0},
\end{equation}
Where
\begin{equation}\label{E}
\begin{aligned}
\bar{E} &= \mathbb{R}^{\dim(\theta)} \times V, \\
V &:= 
\left\{
\Delta \lambda \in \mathbb{R}^{J+T} \;\middle|\;
\begin{array}{l}
\Delta \lambda_g^j \geq 0 \quad \text{for } j \in [1, J] \text{ with } g_j(D, \bar{\theta}) = 0,\ \bar{\lambda}_g^j = 0, \\
\Delta \lambda_g^j = 0 \quad \text{for } j \in [1, J] \text{ with } g_j(D, \bar{\theta}) < 0.
\end{array}
\right\}.
\end{aligned}
\end{equation}
By solving the VI~(\ref{eq:auxiliary VI}), we obtain the changes in the solution and Lagrange multipliers, i.e., $\Delta \theta$ and $\Delta \lambda$. The updated solution after data removal can then be approximated by
$
\theta_{-D^r} \approx \bar{\theta}+\Delta \theta.
$ In the following Section~\ref{sec:3.3}, we reformulate the VI~(\ref{eq:auxiliary VI}) as a quadratic program to facilitate numerical optimization in the context of machine learning-based TSEP.

\section{Machine Unlearning for ML-based TSEP Models}\label{sec:3}
Increasingly, TSEP models are formulated as ML or DL problems. This section instantiates the objective function and constraints of the general constrained learning problem~(\ref{eq:general}) in the context of TSEP, and presents a numerical approach for the machine unlearning method introduced in Section~\ref{sec:2}.

\subsection{ML-Based TSEP Models}
In practical applications, especially with large-scale or high-dimensional data, TSEP models are learned directly from data using empirical loss minimization approaches. The objective function is typically formulated as the sum of individual data losses, i.e., $\sum_1^N \ell(z_i,\theta)$. For example, $\ell(z_i,\theta)=\|f_\theta(x_i,t_i)-z_i\|_2$ measures the discrepancy between the estimated traffic state produced by an ML model $f_\theta$ and the ground truth observation $z_i$ at location $x_i$ and time $t_i$.

The types of constraints in TSEP can vary widely. They may encode domain-specific knowledge (e.g., traffic flow theory) or serve to desirable properties such as fairness or robustness. For example, \cite{xia2025fairtp} introduced region-based static fairness to address the prediction performance gaps between regions caused by uneven sensor deployment. \cite{huang2022physics} introduced the LWR residual error as a physical loss to balance predictive accuracy with theoretical consistency. We briefly summarize the applications and purposes of these constraints in Table~\ref{tab:tsep-constraints}. Note that TSEP models often contain "nominal" constraints, such as flow conservation or nonnegativity, which are omitted from Table~\ref{tab:tsep-constraints} for brevity. Specifically, some research works adopt the regularization techniques to embed knowledge or properties. We treat them as penalty-based relaxations of constrained optimization.

\begin{table}[ht]
\caption{Examples of constraints in ML-based TSEP models.}
\centering
\footnotesize
\begin{tabularx}{\textwidth}{@{}M{3.2cm} L L L@{}}
\toprule
\textbf{Constraint Source} 
& \textbf{Application} 
& \textbf{Constraint Formulation} 
& \textbf{Aim of Constraint} \\
\midrule

\textbf{ML model–driven} 
& Classify vehicle type using SVM~\citep{sun2013vehicle}
& Geometric margin of at least 1 for each training point 
& Ensure a minimum margin for better generalization \\
\midrule

& Traffic density field reconstruction using partially observed data~\citep{huang2022physics,shi2021physics}
& Residual value of the PDE in traffic flow model equals to zero 
& Enforce that the estimated density respects the Greenshields-based LWR \\
\cline{2-4}

\textbf{Domain Knowledge or Physical Models}
& Estimate high-resolution speed fields from sparse probe data~\citep{thodi2022incorporating}
& Limit the model to use information only from directions in which traffic waves realistically propagate
& Characterize the wave propagation \\
\cline{2-4}

& Vehicle longitudinal trajectory prediction on highways~\citep{geng2023physics}
& The predicted trajectory coordinates must follow the Intelligent Driver Model 
& Model the Car Following (CF) behavior of the subject vehicle relative to its leader \\
\midrule

\textbf{Stability-motivated} 
& Network-level speed forecasting~\citep{cui2019traffic}
& L1 norm/L2 norm constraints on the weights of graph neural network
& Reduce model complexity, prevent overfitting \\
\midrule

\textbf{Fairness-motivated} 
& Traffic speed estimation under imbalanced sensor distribution~\citep{xia2025fairtp}
& Disparities in predictive performance between spatial regions 
& Reduce variance in predictions across areas with uneven sensor coverage \\
\bottomrule
\end{tabularx}
\label{tab:tsep-constraints}
\end{table}

To encompass the diverse cases in Table~\ref{tab:tsep-constraints}, we customize the constrained learning problem~(\ref{eq:general}) as the ML-TSEP~(\ref{orginal problem}). The objective function $g_0(D,\theta)$ is expressed as the sum of individual loss terms over the dataset. Since each data point $z_i$ may contribute to one or more constraints, and not every constraint depends on all data points, we define the inequality constraints as $g_j\left(z_{I_j}, \theta\right) \leq 0$ and equality constraints as $h_t\left(z_{I_t}, \theta\right)=0$. Here, $z_{I_j} \subset D$ and $z_{I_t} \subset D$ denote subsets of data points associated with the $j$-th inequality and $t$-th equality constraint, respectively. 
\begin{equation}\label{orginal problem}
\begin{aligned}
\bar{\theta} = \arg \min_\theta & \sum_{i=1}^N \ell\left(z_i, \theta\right)\\
\text{s.t.} \quad 
& g_j\left(z_{I_j}, \theta\right) \leq 0, \quad \forall j \in \{1, \ldots, {J}\}, \\
& h_{t}\left(z_{I_t}, \theta\right) = 0, \quad \forall t \in \{1, \ldots, T\}.
\end{aligned}
\end{equation}
$J$ and $T$ denote the total numbers of inequality and equality constraints, respectively. We focus on removing one data point upon receiving the data deletion request. Without loss of generality, we assume that the deletion request involves only the last data point $z_N$, and $z_N$ contributes exclusively to the  \( J \)-th inequality constraint and the \( T \)-th equality constraint. A general case where multiple data points are removed is provided in Appendix~\ref{Multi-Constraint appendix}.

After removing $z_N$, the ML-TSEP~(\ref{orginal problem}) reduces to the following gold standard ML-TSEP:
\begin{equation}\label{gold standard model}
\begin{aligned}
\bar{\theta}_{-z_N} = \arg \min_\theta &\sum_{i=1}^{N-1} \ell\left(z_i, \theta\right) \\
\text{s.t.} \quad 
& g_j\left(z_{I_j}, \theta\right) \leq 0, \quad \forall j \in \{1, \ldots, J-1\}, \\
& g_J\left(z_{I_J} \setminus \{z_N\}, \theta\right) \leq 0, \\
& h_t\left(z_{I_t}, \theta\right) = 0, \quad \forall t \in \{1, \ldots, T-1\}, \\
& h_T\left(z_{I_T} \setminus \{z_N\}, \theta\right) = 0.
\end{aligned}
\end{equation}
If a constraint only involves $z_{N}$, it will not be enforced. The challenge is to approximate the updated solution $\bar{\theta}_{-z_N}$ based on $\bar{\theta}$ without retraining from scratch while ensuring it satisfies the constraints associated with the remaining data points. 



\subsection{Data Weighed ML-TSEP Model}
To efficiently address the influence of $z_N$, we propose incorporating the constraints of $z_N$ directly into the objective function using a \textit{penalty} function. We call $\phi(C, t)$ a penalty function if it is continuous, non-negative, satisfies $\phi(C, t)=0$ for $t \leq 0$, and is strictly increasing with respect to both $C>0$ and $t>0$. For example, a penalty function would be $
\phi(C, t)= \begin{cases}0 & \text { for } t<0 \\ C t^3 & \text { for } t \geq 0\end{cases}.
$

The original ML-TSEP~(\ref{orginal problem}) is reformulated as:
\begin{equation}\label{penalty problem}
\begin{aligned}
\bar{\theta}_{} = \arg \min_\theta &\sum_{i=1}^{N} \ell(z_i, \theta) + \phi\left(C_g, g_J(z_{I_J}, \theta)\right) + \phi\left(C_h, h_T(z_{I_T}, \theta)\right)+\phi\left(C_h, -h_T(z_{I_T}, \theta)\right)\\
\text{s.t.} \quad 
& g_j(z_{I_j}, \theta) \leq 0, \quad \forall j \in \{1, \ldots, J-1\}, \\
& h_t(z_{I_t}, \theta) = 0, \quad \forall t \in \{1, \ldots, T-1\}.
\end{aligned}
\end{equation}
Where $C_h, C_g$ are the penalty constants that penalize violations of the $z_N$-related constraints.

Following the data weighted constrained learning~(\ref{eq:data weighted tsep}), we introduce a small perturbation $\eta_N$ to weight the terms associated with $z_N$, allowing us to remove the influence of $z_N$ by decreasing the weight $\eta_N$ gradually. The data weighted ML-TSEP is formulated as follows:
\begin{equation}\label{weighted penalty problem}
\begin{aligned}
{\theta}(\eta_N) =  \arg \min_\theta &\bigg\{ 
\sum_{i=1}^{N-1} \ell(z_i, \theta) 
+ \eta_N \cdot \ell(z_N, \theta) 
+ \phi\left(C_g, g_J(z_{I_J} \setminus \{z_N\}, \eta_N \cdot z_N, \theta)\right) \\
& \quad + \phi\left(C_h, h_T(z_{I_T} \setminus \{z_N\}, \eta_N\cdot z_N, \theta)\right) 
+ \phi\left(C_h, -h_T(z_{I_T} \setminus \{z_N\}, \eta_N\cdot z_N, \theta)\right)
\bigg\}\\
\text{s.t.} \quad 
& g_j(z_{I_j}, \theta) \leq 0, \quad \forall j \in \{1, \ldots, J-1\}, \\
& h_t(z_{I_t}, \theta) = 0, \quad \forall t \in \{1, \ldots, T-1\}.
\end{aligned}
\end{equation}
Here $\eta_N$ acts as a tunable parameter that controls the contribution of the terms associated with \( z_N \) in the optimization problem. When $\eta_N = 1$, the data weighted ML-TSEP~(\ref{weighted penalty problem}) reduces to the original ML-TSEP~(\ref{orginal problem}), as $z_N$ is fully included with its original weight. When $\eta_N = 0$, the terms related to $z_N$ are completely removed, and the problem~(\ref{weighted penalty problem}) is equivalent to the gold standard ML-TSEP~(\ref{gold standard model}). 
\subsection{Auxiliary Problem}\label{sec:3.3}
Following Section~\ref{sec:2.1}, conducting the sensitivity of $\theta$ w.r.t. $\eta_N$ ultimately reduces to solving the linearized VI~(\ref{eq:auxiliary VI}). However, the VI formulation is not well-suited for a direct numerical algorithm and thus requires further reformulation. Next, we transfer the VI~(\ref{eq:auxiliary VI}) to a quadratic program, referred to as the \textit{Auxiliary Problem}, to solve the $\Delta \theta$. In the appendix~\ref{aux appendix}, we prove that the optimality condition of the auxiliary problem is exactly the VI~(\ref{eq:auxiliary VI}).

Denoting the Lagrangian function of data weighted ML-TSEP~(\ref{weighted penalty problem}) as $L(  \eta_N,\theta, \lambda)$, we define $z=\left[z_1,z_2,\dots,z_N\right]$ to represent the data points and $\lambda=\left[\lambda_g, \lambda_h\right]$ as the vector of Lagrange multipliers.
Here, $\lambda_g=\left[\lambda_g^j\right]$ corresponds to the multipliers associated with the inequality constraints $g_j(z_{I_j}, \theta) \leq 0$, and $\lambda_h=\left[\lambda_h^t\right]$ corresponds to the multipliers associated with the equality constraints $h_t(z_{I_t}, \theta)$. The Lagrangian function is given by:
\begin{equation}\label{mltesp la}
\begin{aligned}
L( \eta_N,\theta, \lambda) &= 
 \sum_{i=1}^{N-1} \ell(z_i, \theta) 
+ \eta_N \cdot \ell(z_N, \theta) 
+ \phi\left(C_g, g_J(z_{I_J} \setminus \{z_N\}, \eta_N \cdot z_N, \theta)\right) \\
& \quad + \phi\left(C_h, h_T(z_{I_T} \setminus \{z_N\}, \eta_N \cdot z_N, \theta)\right) 
+ \phi\left(C_h, -h_T(z_{I_T} \setminus \{z_N\}, \eta_N \cdot z_N, \theta)\right) \\
& \quad + \sum_{j=1}^{J-1} \lambda_g^j g_j\left(z_{I_j}, \theta\right) 
+ \sum_{t=1}^{T-1} \lambda_h^t h_t\left(z_{I_t}, \theta\right).
\end{aligned}
\end{equation}
Given the (\ref{orginal problem}) with its optimal solution $\bar{\theta}$ and the corresponding Lagrangian multiplier $\bar{\lambda}$. We have $\left[\bar{\theta},\bar{\lambda}\right]=\left[\theta(\eta_N=1),\lambda(\eta_N=1)\right]$. The auxiliary problem~(\ref{Aux problem}) is designed to measure $\Delta \theta$, as $\eta_N$ moves away from $\bar{\eta}_N=1$ along the direction of $q=-1$.
\begin{equation}\label{Aux problem}
\begin{aligned}
\min_{\Delta \theta} & \quad \bar{g}_0(\Delta \theta) + \langle \nabla_{\theta \eta_N}^2{L}(\bar{\eta}_N,\bar{\theta},  \bar{\lambda})q, \Delta \theta \rangle \\
\text{s.t.} & \quad 
\bar{g}_j(\Delta \theta, z_{I_j}) \begin{cases}
= 0 & \text{for } i \in I \setminus I_0, \\
\leq 0 & \text{for } i \in I_0, \\
\text{free} & \text{for } i \in I_1,
\end{cases} \\
& \quad \bar{h}_t(\Delta \theta, z_{I_t}) = 0, \quad t \in \{1, 2, \dots, T-1\}.
\end{aligned}
\end{equation}
where the three index sets $I, I_0$, and $I_1$ are defined as:
\begin{equation*}
\begin{aligned}
I &= \left\{j \in [1, J-1] \mid g_j( z_{I_j},\bar{\theta}) = 0 \right\}, \\
I_0 &= \left\{j \in [1, J-1] \mid g_j(z_{I_j},\bar{\theta}) = 0 \text{ and } \bar{\lambda}_g^j = 0 \right\} \subset I, \\
I_1 &= \left\{j \in [1, J-1] \mid g_j(z_{I_j},\bar{\theta}) < 0 \right\}.
\end{aligned}
\end{equation*}
Additionally, $\bar{g}_0(\Delta \theta)$ is the second order expansion of ${L}(z, \theta, w, \lambda)$:
\begin{equation}
\bar{g}_0(\Delta \theta)=L(\bar{\eta},  \bar{\theta}, \bar{\lambda})+\left\langle\nabla_\theta L(\bar{\eta}, \bar{\theta}, \bar{\lambda}), \Delta \theta\right\rangle+\frac{1}{2}\left\langle\Delta \theta, \nabla_{\theta \theta}^2 L(\bar{\eta}, \bar{\theta},  \bar{\lambda}) \Delta \theta\right\rangle,
\end{equation}
and $\bar{g}_j(\Delta \theta, z_{I_j}),\bar{h}_t(\Delta \theta, z_{I_t})$ are the first-order expansion of constraints:
\begin{equation}
\bar{g}_j(\Delta \theta, z_{I_j})=g_j(z_{I_j},\bar{\theta})+\left\langle\nabla_{\theta}g_j(z_{I_j},\bar{\theta}),\Delta \theta\right\rangle, \bar{h}_t(\Delta \theta, z_{I_t})=h_t(z_{I_t},\bar{\theta})+\left\langle\nabla_{\theta}h_t(z_{I_t},\bar{\theta}),\Delta \theta\right\rangle.
\end{equation}

By solving the auxiliary problem~(\ref{Aux problem}) for $\Delta \theta$, we can determine how $\theta$ changes from $\bar{\theta}$ when $\eta_N$ moves from $\bar{\eta}_N$ along the direction of $q$. Also note that problem~(\ref{Aux problem}) is a quadratic optimization problem, which is normally much easier to solve than retraining the original ML-TSEP model.
\subsection{Some Remarks}

When TSEP has no constraints, our method can be simplified. For example, assume the data weighted ML-TSEP~(\ref{weighted penalty problem}) model is 
\begin{equation}
\bar{\theta}=\arg \min _\theta \sum_{i=1}^{N-1} \ell\left(z_i, \theta\right)+\eta_N \cdot \nabla_\theta\ell (z_N, \theta).\end{equation}
The optimality condition reduces to 
\begin{equation}
\sum_{i=1}^{N-1}\nabla_{{\theta}} \ell\left(z_i, \bar{\theta}\right)+\eta_N \cdot \ell (z_N,\bar{\theta})=\mathbf{0}.\end{equation}
The influence of data point $z_N$ on the solution $\bar{\theta}$ can be measured by its derivative, which is derived using the Dini implicit function theorem~\citep{krantz2002implicit}:
\begin{equation}
\left. \frac{\partial \theta}{\partial \eta_N} \right|_{\eta_N = 1}
= -\left[ \sum_{i=1}^N \nabla_{\theta}^2 \ell\left(z_i, \bar{\theta} \right) \right]^{-1} 
\nabla_{\theta} \ell\left(z_N, \bar{\theta} \right).
\end{equation}
The $\Delta \theta$ can be estimated by:
\begin{equation}
    \Delta \theta = \theta(\eta_N=0)-\theta(\eta_N=1)\approx \frac{\partial \theta}{\partial \eta_N}|_{\eta_N = 1} \cdot (-1).
\end{equation}

When constraints are introduced, the optimality condition can be characterized by the Karush-Kuhn-Tucker (KKT) conditions, rendering the Dini implicit function theorem inapplicable. Moreover, the Dini implicit function theorem requires the loss function to have a non-singular Hessian matrix, a condition that is often violated in deep learning scenarios due to the non-convex nature of neural networks. By leveraging the generalized implicit function theorem proposed by~\citet{dontchev2009implicit}, our proposed methods can address these issues and thus extend the influence function to constrained learning problems. This extension further enables the simultaneous removal of multiple data points; see Appendix~\ref{Multi-Constraint appendix}.


\section{Application} \label{sec:4}
We implement our machine unlearning model and algorithm on two TSEP applications using trajectory data. The first one is an SVM-based vehicle classification model that utilizes GPS trajectories to distinguish between passenger cars and trucks. The second application employs a physics-informed neural network (PINN) to reconstruct vehicle velocity fields using Next Generation SIMulation (NGSIM) dataset.

Given that trajectory data often contains sensitive personal information—such as home and work locations, travel habits, and frequently visited places—many jurisdictions mandate that organizations allow users to opt out of data collection or request data erasure. Additionally, malicious actors can inject poisoned trajectory data to manipulate model predictions, potentially fabricating congestion patterns or altering autonomous vehicle behavior. When any of these scenarios occurs, the sensitive or poisoned data points may need to be removed, leading to the machine unlearning problem as discussed here. In this case, our approach ensures compliance with privacy regulations and enhances model integrity against adversarial attacks in a computationally efficient manner.

\subsection{Machine Unlearning of SVM}\label{sec:4.1}
We evaluate our method on an SVM-based vehicle classification model. The classification task utilizes acceleration and deceleration features derived from real-world vehicle GPS trajectories to divide vehicles into passenger cars or trucks. Details about the model and dataset can be found in~\citep{sun2013vehicle}.
\subsubsection{Data Weighted SVM}
Denoting the data as $z_i=\left[x_i, y_i\right]$, where $i \in\{1,2, \ldots, N\}, x_i$ represents the feature vector, and $y_i \in\{-1,1\}$ is the label, the soft-margin SVM model can be formulated as:
\begin{equation}\label{SVM}
\begin{aligned}
\min_{\theta=\{w, b, \xi\}} \quad  &\frac{1}{2}\|w\|^2+C \sum_{i=1}^N \xi_i \\
\textrm { s.t.     }\quad &y_i\left(\eta \cdot x_i+b\right) \geqslant 1-\xi_i,\\
&\xi_i \geqslant 0.\\
&i=1, \ldots, N
\end{aligned}
\end{equation}
where $C>0$ is the regularization parameter penalizing the slack variables $\xi_i$.

In this section, we assume we aim to remove the last data $z_N$. Following the data weighted ML-TSEP~(\ref{weighted penalty problem}), the data weighted SVM model is:
\begin{equation}\label{weighted SVM}
\begin{aligned}
\min_{\theta=\{w,b,\xi\}} \quad &\frac{1}{2}\|{w}\|^2 + C\sum_{i=1}^{N-1} \xi_i+\eta_N \cdot C \max\left[0,1-y_N\left({w}^\top x_N + b\right)\right] \\
\textrm { s.t       } \quad &y_i\left({w}^\top x_i + b\right) \geq 1 - \xi_i, \\
&\xi_i \geq 0. \\
&i=1, \ldots, N-1.
\end{aligned}
\end{equation}

It is straightforward to verify that model (\ref{weighted SVM}) is equivalent to model (\ref{SVM}) when $\eta_N=1$, and is equivalent to the SVM model trained on the remaining $N-1$ data points (gold standard model) when $\eta_N=0$. By estimating how the solution changes as we decrease $\eta_N$ from 1 to 0, we can fulfill the data removal request. The sensitivity analysis of solution ${\theta=\{w,b,\xi\}}$ w.r.t. $\eta_N$ can be found in the appendix~\ref{svm appendix}. 

\subsection{Results}
\begin{figure}
    \centering
    \includegraphics[width=0.7\linewidth]{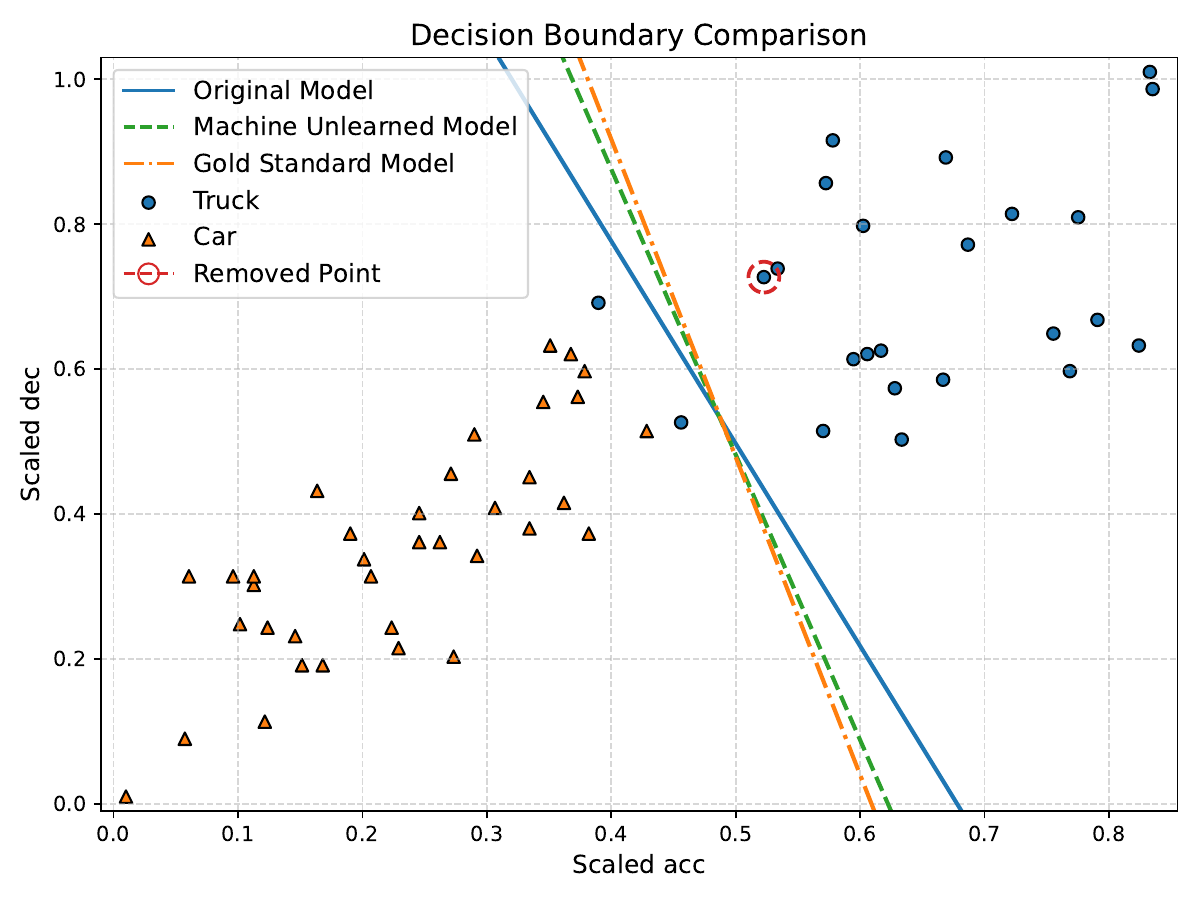}
    \caption{Comparison of Decision Boundaries: Original SVM, Unlearned Model, and Gold Standard Model}
    \label{fig:SVM Simulation}
\end{figure}

We first retrain the model after data removal to obtain the gold standard model (ground truth). Subsequently, we calculate the solution change using the machine unlearning approach described in Section~\ref{sec:3}. Table \ref{tab:comparison} summarizes the parameters and classification accuracy of the original model, the machine unlearned model, and the gold standard model. Figure \ref{fig:SVM Simulation} illustrates the decision boundaries of the different models. First, the SVM solution undergoes significant changes due to the removal of just one data point. This underscores the importance of developing efficient machine unlearning algorithms to handle the data removal request. Second, both the plotted decision boundaries and the numerical comparisons underscore the accuracy of the proposed machine unlearning method in closely approximating the gold standard solution. These results validate the ability of the proposed unlearning model to maintain model accuracy while effectively removing the influence of specific data points. 
\begin{table}[h!]
\centering
\caption{Comparison of SVM Models After Unlearning}
\begin{tabular}{lccc|cc}
\toprule
\textbf{Model}            & \(\mathbf{w_1}\) & \(\mathbf{w_2}\) & \(\mathbf{b}\) & \textbf{Train Accuracy} & \textbf{Test Accuracy} \\ 
\midrule
Original Model     & -7.4230          & -2.6563          & 5.0319         & 0.9667                  & 0.9333                 \\
Machine unlearned Model          & -7.3818          & -1.8735          & 4.5935         & 0.9667                  & 0.9500                 \\
Gold Standard Model & -7.3402          & -1.6735          & 4.4707         & 0.9661                  & 0.9500                 \\ 
\bottomrule
\end{tabular}
\label{tab:comparison}
\end{table}

\subsection{Machine Unlearning for PINN-Based Traffic State Estimation}
\subsubsection{PINN-Based Traffic State Estimation}
The PINN model we test here is based on~\cite{shi2021physics},  which reconstructs the velocity field using the observed vehicle velocity data extracted from the NGSIM dataset. By integrating physical constraints with data-driven learning, the model ensures that the reconstructed velocity field adheres to the LWR model.

\textit{Data Description:} We extract vehicle trajectory data from the NGSIM dataset for a road segment on I-80 in Emeryville, California, which spans approximately $L=1600$ feet. The set of vehicle trajectories is represented by $\mathscr{T}=\left\{\left(x_i, t_i, v_i\right)\right\}$, where each entry represents a vehicle's position $x_i$, timestamp $t_i$, and speed $v_i$. The trajectory to be removed is $\mathcal{T}^r$. The total duration of the data is $T=15$ minutes.

\begin{figure}
    \centering
    \includegraphics[width=0.6\linewidth]{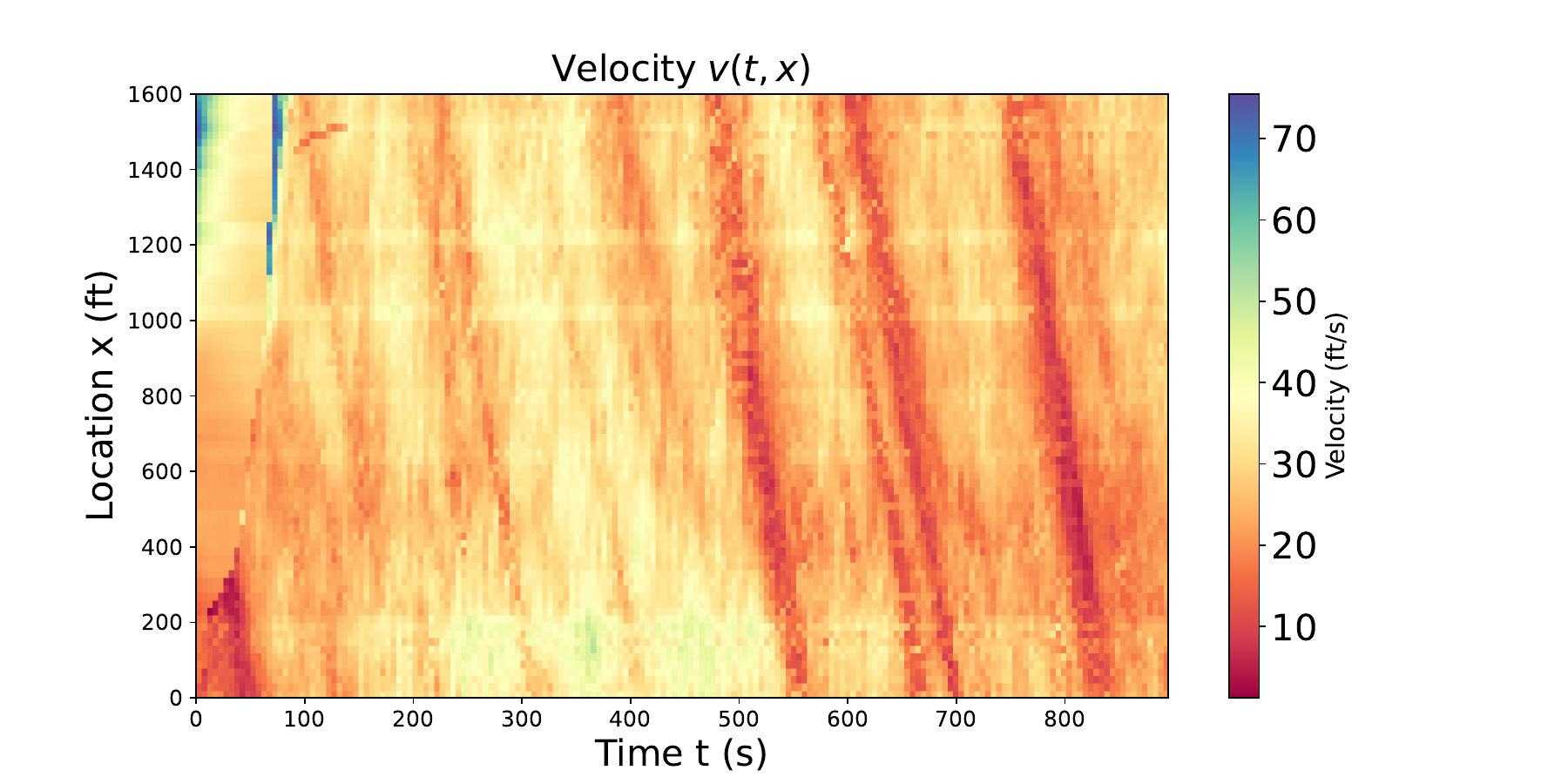}
    \caption{Visualization of the vehicle velocity field on I-80 from NGSIM}
    \label{fig:velocity visualization}
\end{figure}

\textit{Data Processing:} Following the data processing in~\cite{huang2022physics}, we construct the velocity field using a binning method with a spatial resolution  $\Delta x$  of 20 feet and a temporal resolution  $\Delta t$  of 5 seconds.  The center of each cell is treated as a grid point.  We denote the spatial-temporal grid by $\Omega = \{(x_p,t_p) | p=1,2,\dots,N_p\}$, where $x_p$ and $t_p$ represent the spatial and temporal positions of the grid points, and $N_p=\frac{L}{\Delta x} \times \frac{T}{\Delta t} $ is the total number of grid points. The average speed of vehicles is computed for each bin, resulting in a velocity field consisting of 180 temporal bins and 80 spatial bins. The mean speed in each bin is computed as: 
\begin{equation}\label{mean speed}
v(x_p,t_p)=\frac{1}{\left|{S}(x_p,t_p)\right|} \sum_{\left(x_i, t_i, v_i\right)  \in {S}(x_p,t_p)} v_i
\end{equation}
where ${S}(x_p,t_p)=\left\{\left(x_i, t_i, v_i\right) \in \mathcal{T} \mid x_i \in\left[x_p-\frac{1}{2}\Delta x, x_p+\frac{1}{2}\Delta x\right], t_i \in\left[t_p-\frac{1}{2}\Delta t, t_p+\frac{1}{2}\Delta t\right]\right\}$ is the set of trajectory data points that fall within the spatial-temporal bin. $\left|{S}(x_p,t_p)\right|$ is the number of data points in the bin. Only a subset of the velocity field (14\% of samples) will be used as \textit{observed data} to train the model, while the remaining values will serve as the ground truth. 
We denote the observed data index as $\mathbf{O} = \{(x_o,t_o) | o=1,2,\dots,N_o\}$. The observed data $v(x,t), (x,t)\in \mathbf{O}$ is used to reconstruct the whole velocity field. We also choose a subset of the grid points $\mathbf{A}=\{(x_a,t_a)|a=1,2,\dots,N_a\}$, as the \textit{auxiliary points} to examine the physical constraints. Figure~\ref{fig:velocity visualization} shows the velocity field on the I-80 freeway from 4:00 p.m. to 4:15 p.m., based on the NGSIM dataset. The presence of shock waves can also be observed.

\textit {PINN Model Details:} We follow the PINN model proposed by~\cite{shi2021physics}. The goal of PINN is to learn a neural network $\hat{v}(x, t; \theta)$, where $(x, t)$ are the inputs and $\theta$ represents the trainable parameters, to approximate $v(x, t)$.  At the same time, $\hat{v}(x, t; \theta)$, as the traffic state estimator, must adhere to the fundamental dynamics of traffic flow. These dynamics are governed by the LWR model. We assume that density $\rho$, flow $q$, and speed $v$ follow Greenshields' fundamental diagram. 
To facilitate our discussion, the PDE of the LWR is given by: 
\begin{equation}
\mathcal{N}(\hat{v}(x,t,\theta), Q(\hat{v}(x,t,\theta)); {\omega})={0}, x \in[0, L], t \in[0, T],\end{equation}
where  $\omega:=\{v_f,\rho_m\}$ denotes the given free-flow speed and jam density, and $Q(\hat{v})$ denotes the intermediate unobserved traffic variables (flow, density) that have some hidden relationship with $\hat{v}$.  


\subsubsection{Data Weighted PINN Model}
The \textbf{original PINN} can be formulated as:

\begin{equation}
\label{original pinn}
\begin{aligned}
\min_\theta \quad & \sum_{(x_o,t_o) \in \mathbf{O}} \left\| \hat{v}(x_o,t_o,\theta) - v(x_o, t_o) \right\|^2_2, \\
\text{s.t.} \quad & 
\mathcal{N}(\hat{v}(x_a,t_a,\theta), Q(\hat{v}(x_a,t_a,\theta)); {\omega}) = 0, 
& \forall (x_a,t_a) \in \mathbf{A}\\
\quad&  v(x_o,t_o)=\frac{1}{\left|{S}(x_o,t_o)\right|} \sum_{\left(x_i, t_i, v_i\right)  \in {S}(x_o,t_o)} v_i. 
\end{aligned}
\end{equation}

We use $\mathscr{T}^r \subset \mathscr{T} $ to denote the trajectory data requested to be removed and define $S^r\left(x_p, t_p\right)=\left\{\left(x_i, t_i, v_i\right) \in \mathcal{T}^r \mid x_i \in\left[x_p-\frac{1}{2}\Delta x, x_p+\frac{1}{2}\Delta x\right], t_i \in\left[t_p-\frac{1}{2} \Delta t, t_p+\frac{1}{2}\Delta t\right]\right\}$ as the set of removed trajectory data that fall within the spatial-temporal cell around $(x_p,t_p)$. The mean speed at $(x_p,t_p)$, defined by~(\ref{mean speed}) is reformulated as:
\begin{equation}
v\left(x_p, t_p,\eta\right)=\frac{1}{\left|S^k\left(x_p, t_p\right)\right|+\eta\left|S^r\left(x_p, t_p\right)\right|}\left(\sum_{\left(x_i, t_i, v_i\right) \in S^k\left(x_p, t_p\right)} v_i+\eta \sum_{\left(x_i, t_i, v_i\right) \in S^r\left(x_p, t_p\right)} v_i\right),
\end{equation}
where $S^k\left(x_p, t_p\right)=S(x_p,t_p)\setminus S^r(x_p,t_p)$. $\eta$ denotes the weight assigned to the trajectory data that needs to be removed.
The  \textbf{Data-weighted PINN model} is formulated as:

\begin{equation}
\label{data weighted pinn}
\begin{aligned}
\min_\theta \quad & \sum_{(x_o,t_o) \in \mathbf{O}} \left\| M(x_o,t_o,\theta) - v_{w}(x_o, t_o) \right\|_2^2, \\
\text{s.t.} \quad & 
\mathcal{N}({M}(x_a,t_a,\theta), Q(M(x_a,t_a,\theta)); {\omega}) = 0, 
\forall (x_a,t_a) \in \mathbf{A}\\
&v_w\left(x_o, t_o\right)=\frac{1}{\left|S^k\left(x_o, t_o\right)\right|+\eta\left|S^r\left(x_o, t_o\right)\right|}\left(\sum_{\left(x_i, t_i, v_i\right) \in S^k\left(x_o, t_o\right)} v_i+\eta \sum_{\left(x_i, t_i, v_i\right) \in S^r\left(x_o, t_o\right)} v_i\right).
\end{aligned}
\end{equation}
It is trivial to see the problem (\ref{data weighted pinn}) is equivalent to the original PINN (\ref{original pinn}) when $\eta=1$.
When $\eta$ shifts from 1 to 0, the problem (\ref{data weighted pinn}) shifts to the PINN trained on the remaining data $\mathscr{T}\setminus\mathscr{T}^r$. 

\subsubsection{Experiment on PINN}
\begin{figure}[htbp]
    \centering
    \begin{subfigure}[t]{0.32\textwidth}
        \includegraphics[width=\linewidth]{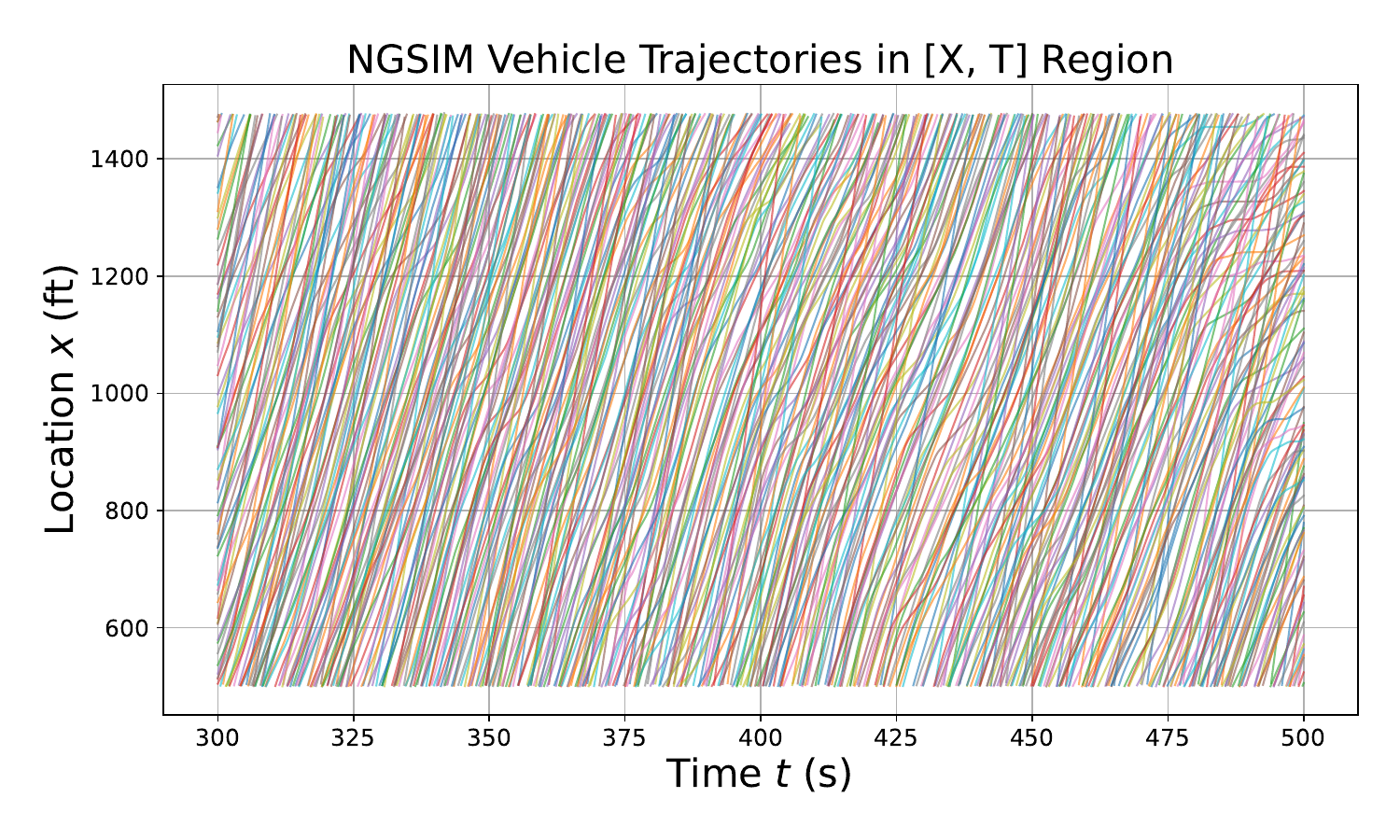}
        \caption{}
        \label{fig:sub1}
    \end{subfigure}
    \hfill
    \begin{subfigure}[t]{0.32\textwidth}
        \includegraphics[width=\linewidth]{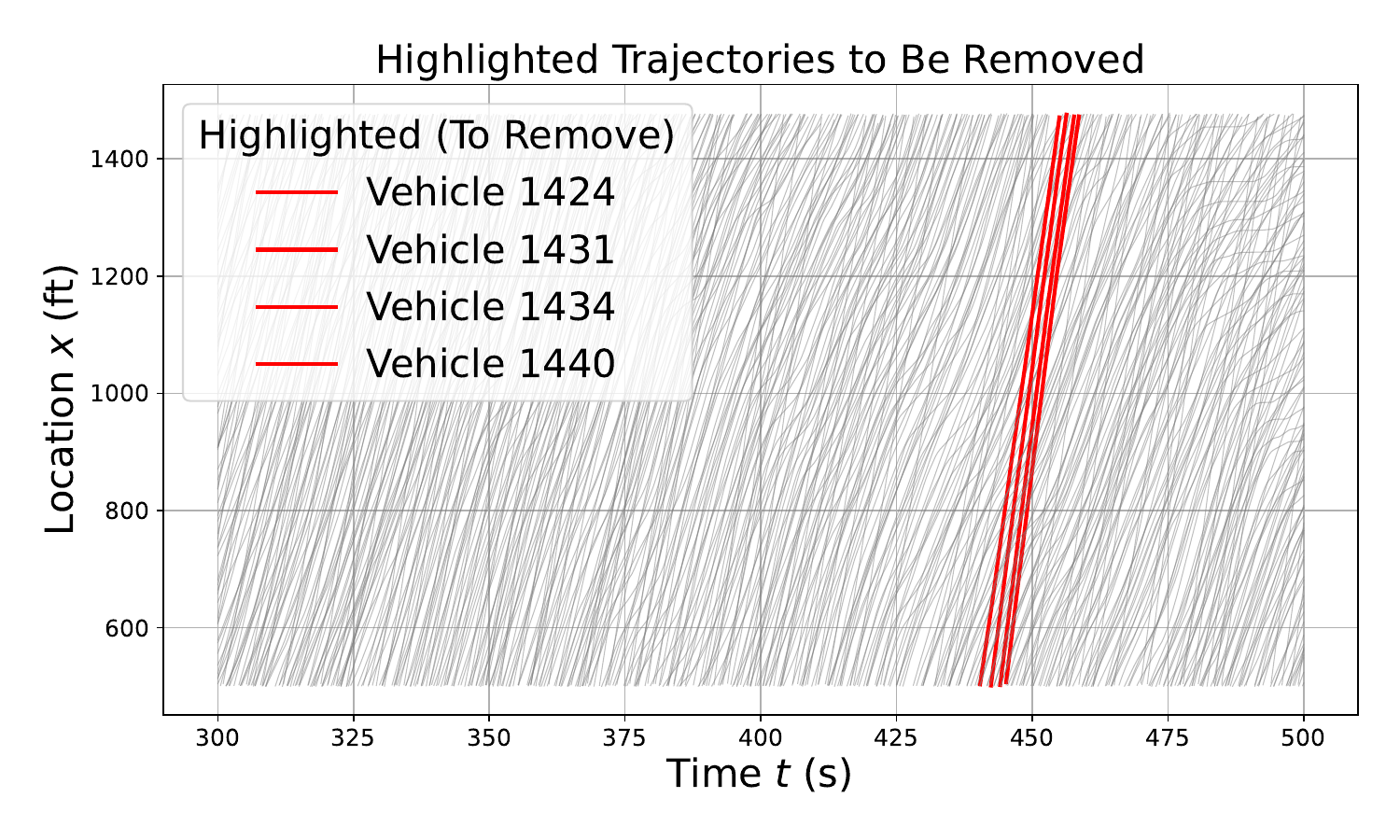}
        \caption{}
        \label{fig:sub2}
    \end{subfigure}
    \hfill
    \begin{subfigure}[t]{0.32\textwidth}
        \includegraphics[width=\linewidth]{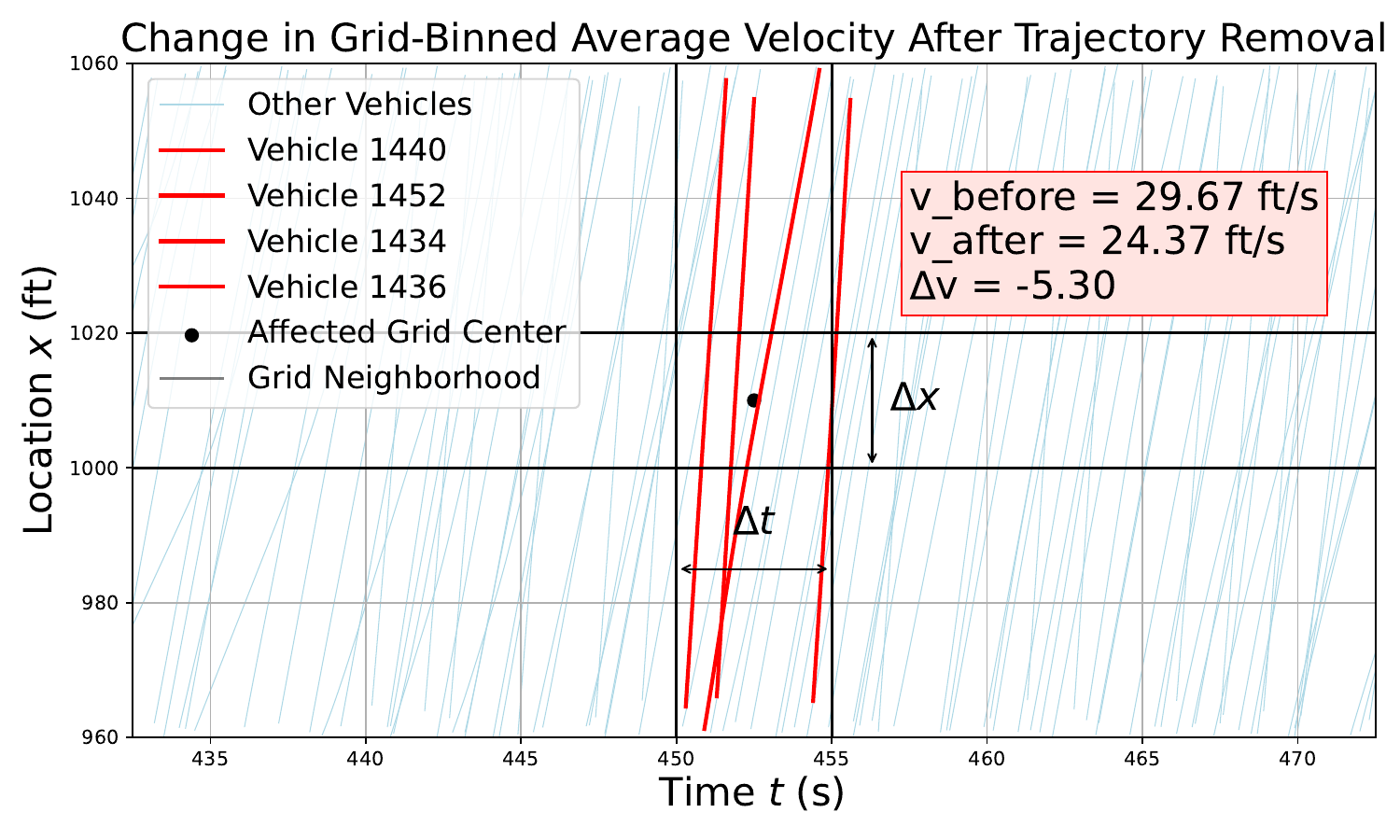}
        \caption{ }
        \label{fig:sub3}
    \end{subfigure}
    \caption{Illustration of trajectory removal and its impact on observed velocity.  (a) All trajectories. (b) Subset of removed trajectories. (c) Change in average velocity in affected spatiotemporal bins due to trajectory removal.}
    \label{fig:three_images}
\end{figure}

 \textit{Experiment setting:} We consider removing 10\% of the training data (about 12,660 trajectories), which results in modifications to the observed average speed in certain spatiotemporal bins. Our machine unlearning method adjusts the original PINN parameters to accommodate this change, and the updated model is then compared to the gold standard model with retraining. Fig.~\ref{fig:three_images} details the data removal workflow. All experiments are conducted on an RTX 4090 GPU. 

Table~\ref{tab:pinnresults} presents the performance of three PINN models--Original, Retrained, and Unlearned--on the residual dataset after data removal. The Retrained Model achieves the best performance across all metrics, with the lowest data loss (MAE: 2.495 ), physics loss (MAE: $1.12 \times 10^{-2}$ ), and relative $L_2$ error ( $14.24 \%$ ). The Unlearned Model shows slightly higher errors than the Retrained Model, but remains significantly better than the Original Model.
Notably, the Unlearned Model requires only 91.77 seconds to update the parameters, which is approximately $3.6 \times$ faster than full retraining, demonstrating the computational advantage of our machine unlearning approach. Note that the PINN used in our experiments are relatively small, consisting of nine layers with approximately 10,000 parameters. Even for such a compact DNN, the unlearned model can save up to three times the training time. For larger neural networks, where the cost of full retraining is significantly higher, our unlearning method is expected to offer even greater time savings.

\begin{table}[ht]
\centering
\caption{Performance Comparison of Different PINN Models on the Residual Dataset After Removal}
\begin{tabular}{lcccc}
\toprule
\textbf{Model} & \makecell{\textbf{MAE} \\ \textbf{(Data Loss)}}& \makecell{\textbf{MAE} \\ \textbf{(Physics Loss)}} & \textbf{Relative $L_2$ Error (\%)} & \textbf{Training Time (s)} \\
\midrule
Original Model    & 4.156 & \({47.1 \times 10^{-2}}\)
 & 21.65 & 327.06 \\
Retrained Model  & 2.495 & \({1.12 \times 10^{-2}}\)
 & 14.24 & 331.50 \\
Unlearned Model  & 2.832 & \({0.37 \times 10^{-2}}\)
 & 16.68 & 91.77 \\
\bottomrule
\end{tabular}
\label{tab:pinnresults}
\end{table}

\begin{figure}
    \centering
    \includegraphics[width=1\linewidth]{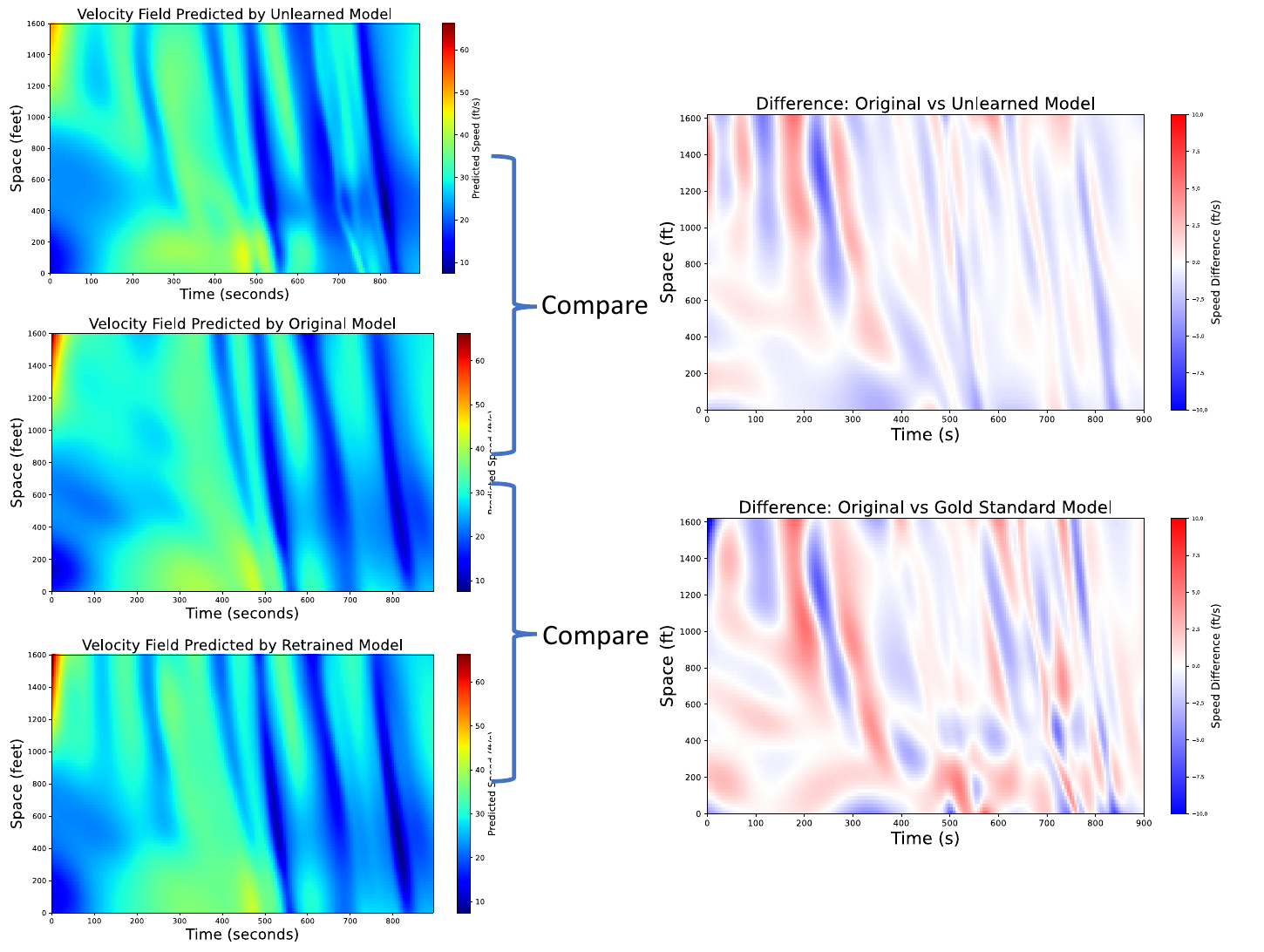}
    \caption{Predicted velocity field differences after model unlearning and retraining (12660 trajectories removed).}
    \label{fig:comparevelocityfield}
\end{figure}

Figure~\ref{fig:comparevelocityfield} illustrates the predicted velocity fields and their differences after removing 10\% of the trajectories. The left column shows the velocity fields predicted by the Unlearned, Original, and Retrained models, respectively. The right column visualizes the absolute differences in predicted speeds: the top figure compares the Unlearned and Original models, while the bottom figure compares the Retrained (gold standard) and Original models. The difference maps reveal that the Unlearned model achieves a comparable correction effect to that of the fully Retrained model, but with significantly lower computational cost. These results confirm the effectiveness of our machine unlearning approach in adapting the model to data removal while maintaining high fidelity in the reconstructed velocity field.

\section{Conclusions and Future Research} \label{sec:5}
This paper proposes a machine unlearning framework tailored to TSEP applications. The goal is to enable TSEP models to forget privacy-sensitive, poisoned, or outdated data without retraining from scratch. Building upon influence function theory, we develop a sensitivity analysis framework to approximate the change in optimal model parameters, $\Delta \theta$, resulting from the removal of one or more data points. To the best of our knowledge, this is the first machine unlearning algorithm specifically designed for constrained learning scenarios.

The proposed method is further instantiated in two representative learning paradigms: (i) support vector machines (SVMs), where inequality constraints encode margin conditions, and (ii) physics-informed neural networks (PINNs), where the learned solution must respect the traffic flow theory. In both cases, our approach significantly reduces computational cost compared to the gold standard model with retraining, while preserving key performance metrics. Overall, this work bridges the gap between machine unlearning on TSEP, robust statistics, and sensitivity analysis, offering a novel perspective on privacy-preserving, safety-aware, and computationally efficient learning pipelines.

This work opens several promising directions for further research. The first is real-time or streaming unlearning.  While we focus on offline unlearning scenarios, future work may extend the proposed methods to handle streaming traffic data, where traffic models must adapt dynamically to forget obsolete or withdrawn data. For this, prior solutions and intermediate optimization states may be reused to facilitate efficient unlearning.  The second direction is to explore the use of machine unlearning to defend against adversarial attacks.  In TSEP, carefully crafted adversarial data injected during training (poisoning attacks) can lead to significant degradation in model performance or even safety critical failures. Existing defense methods often rely on adversarial training or robust optimization, which may be computationally expensive and hard to generalize. Machine unlearning offers an alternative line of defense: once adversarial or suspicious data is detected, the model can “forget” the influence of these harmful inputs without full retraining. This reactive approach may complement existing defenses and be particularly useful in scenarios where attacks are detected post hoc or evolve over time. 

\newpage
\appendix
\section{Mathematical details}
\subsection{Notation Summary}

\begin{table}[h]
\centering
\caption{Summary of Notations}
\label{tab:notation-summary}
\begin{tabularx}{\textwidth}{@{}lX@{}}
\toprule
\textbf{Symbol} & \textbf{Description} \\
\midrule
$D = \{z_1, z_2, \dots, z_N\}$ & Full dataset consisting of $N$ data points. \\
$z_i$ & $i$-th data point, may include features and label. \\
$D^r$ & Subset of data to be removed. \\
$\theta$ & Model parameters. \\
$\bar{\theta}$ & Optimal solution before data removal. \\
$\bar{\theta}_{-D^r}$ & Optimal solution after removing data $D^r$. \\
$\eta = [\eta_1, \dots, \eta_N]$ & Data weight vector. \\
$\eta^r$ & Data weight vector after removing $D^r$ (i.e., some entries are 0). \\
$\ell(z_i, \theta)$ & Loss function for data point $z_i$. \\
$g_j(\cdot)$ & $j$-th inequality constraint. \\
$h_t(\cdot)$ & $t$-th equality constraint. \\
$\lambda = (\lambda_g, \lambda_h)$ & Lagrange multipliers for constraints. \\
$\mathcal{L}(\eta, \theta, \lambda)$ & Lagrangian of the weighted optimization problem. \\
$f(\eta, \theta, \lambda)$ & Stationarity residual used in variational inequality (VI). \\
$N_E(\cdot)$ & Normal cone to the feasible region $E$. \\
$\Delta \theta$, $\Delta \lambda$ & Change in solution and multipliers due to data removal. \\
$\mathcal{N}(\cdot)$ & PDE residual of the LWR model in PINNs. \\
$v(x,t)$ & Observed average velocity. \\
$\hat{v}(x,t;\theta)$ & Predicted velocity field by the PINN. \\
$S^r(x,t)$ & Removed trajectory points in the spatiotemporal bin around $(x,t)$. \\
$v(x,t;\eta)$ & Weighted average velocity after down-weighting $S^r$. \\
$\phi(C, t)$ & Penalty function applied to constraint violations. \\
$\mathbf{O}, \mathbf{A}$ & Index sets for observed data and auxiliary PDE points. \\
\bottomrule
\end{tabularx}
\end{table}

\subsection{Normal Cone Definition}\label{normal cone appendix}

Let ${E} \subseteq \mathbb{R}^d$ be a closed convex set, and let $\theta \in {E}$. The normal cone to ${E}$ at point $\theta$, denoted by $N_{E}(x)$, is defined as~\citep{dontchev2009implicit}

$$
N_{{E}}(\theta)=\left\{{u} \in \mathbb{R}^d \mid\langle {u}, \theta^\prime-\theta\rangle \leq 0, \forall \theta^\prime \in E\right\} .
$$

\begin{figure}
    \centering
    \includegraphics[width=0.4\linewidth]{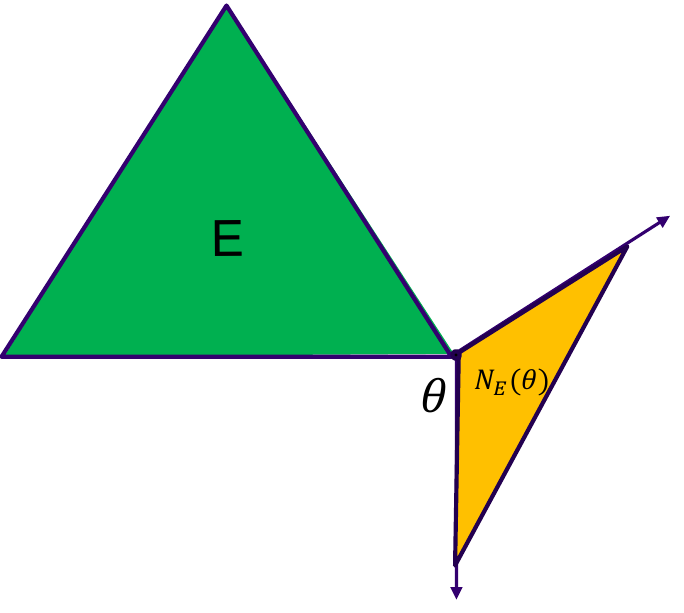}
    \caption{Example of Normal cone}
    \label{fig:normal cone}
\end{figure}

This set contains all vectors that form non-acute angles with every feasible direction from $\theta$ within $E$. Intuitively, it characterizes the directions of non-descent at the boundary of the feasible set. An example is provided in Figure ~\ref{fig:normal cone}.

\subsection{Optimality Condition of Auxiliary Problem }\label{aux appendix}

We present the derivation of the variational inequality (VI) optimality condition for the auxiliary problem~(\ref{Aux problem}).

Let ${L}(\eta_N,\theta,  \lambda)$ denote the Lagrangian of the ML-TSEP problem defined in~(\ref{mltesp la}).  We use $L$ to denote the Lagrangian ${L}( \bar{\eta},\bar{\theta},  \bar{\lambda})$ for brevity in Appendix~\ref{aux appendix}.

Following~(\ref{eq:auxiliary VI}), the auxiliary VI of ML-TSEP is: 
\begin{equation}\label{33}
\nabla_{\eta_N} f(\bar{\eta}, \bar{\theta}, \bar{\lambda})\Delta \eta_N+\nabla_{(\theta, \lambda)} f(\bar{\eta}, \bar{\theta}, \bar{\lambda})[\Delta \theta ; \Delta \lambda]+N_{\bar{E}}(\Delta \theta, \Delta \lambda) \ni \mathbf{0},
\end{equation}
where  
\begin{equation}
f(\eta, \theta, \lambda)=\left(\nabla_\theta L,-\nabla_\lambda L\right)^{\top}
\end{equation}
and 
\begin{equation}
\begin{aligned}
& \bar{E}=\mathbb{R}^{\operatorname{dim}(\theta)} \times V, \\
& V:=\left\{\Delta \lambda \in \mathbb{R}^{J+T-2} \left\lvert\, \begin{array}{ll}
\Delta \lambda_g^j \geq 0 & \text { for } j \in[1, J-1] \text { with } g_j(D, \bar{\theta})=0, \bar{\lambda}_g^j=0, \\
\Delta \lambda_g^j=0 & \text { for } j \in[1, J-1] \text { with } g_j(D, \bar{\theta})<0 .
\end{array}\right.\right\} .
\end{aligned}
\end{equation}
The VI~(\ref{33}) can be translated to the requirements that:
\begin{equation}\label{requirements}
    \langle\nabla_\theta^2L,\Delta \theta\rangle+\langle \nabla_{\theta\lambda}L, \Delta \lambda\rangle+\langle\nabla_{\theta\eta_N} L,\Delta\eta_N\rangle=0,
\end{equation}
where $\Delta \lambda_g^j \geq 0$ if $j \in I_0$ having $\bar{g}_j\left(\Delta \theta, z_{I_j}\right)=0$. All other $\Delta \lambda_g^j$ are zero.

We define the lagragian function of auxiliary problem~(\ref{Aux problem}):
\begin{equation}
\begin{aligned}
{L}_{\mathrm{aux}}(\Delta \theta, \gamma) =
&{L}
+ \left\langle \nabla_\theta {L}, \Delta \theta \right\rangle + \frac{1}{2} \left\langle \Delta \theta,\ \nabla^2_{\theta \theta} {L} \cdot \Delta \theta \right\rangle \\
&+ \left\langle \nabla^2_{\theta \eta_N} {L} \cdot q,\ \Delta \theta \right\rangle + \sum_{j=1}^{J-1} \gamma_g^j g_j(z_{I_j}, \bar{\theta})+\sum_{j=1}^{J-1} \gamma_g^j\left\langle \nabla_\theta g_j(z_{I_j}, \bar{\theta}),\ \Delta \theta \right\rangle
+ \sum_{t=1}^{T-1} \gamma_h^t \left\langle \nabla_\theta h_t(z_{I_t}, \bar{\theta}),\ \Delta \theta \right\rangle,
\end{aligned}
\end{equation}
where $\gamma=[\gamma_g^j,\gamma_h^t]$ are the Lagrangian multipliers. We have used the fact that $h_t(\bar{\theta}) = 0$, so constant terms in the constraint expansions vanish. 
The optimality condition of the auxiliary problem is:
\begin{equation}\label{35}
    \nabla_{\Delta\theta} {L}_{\mathrm{aux}}(\Delta\theta,\gamma) = 0,
\end{equation}
Note that $\nabla_\theta L=0$ due to KKT of ML-TSEP~(\ref{orginal problem}) and $\nabla_\lambda L=[g_j\left(z_{I_j}, \theta\right),h_t\left(z_{I_t}, \theta\right)]$.  By substituting $q=\Delta \eta_N$ to~(\ref{35}), we have
\begin{align}\label{36}
    \nabla_{\Delta \theta} L_{\text {aux }}(\Delta \theta, \gamma) &= \left\langle\nabla_\theta^2 L, \Delta \theta\right\rangle+\nabla_{\theta \eta_N}^2 L \cdot q+\sum_{j=1}^{J-1} \gamma_g^j \nabla_\theta g_j\left(z_{I_j}, \bar{\theta}\right)+ \sum_{t=1}^{T-1} \gamma_h^t\nabla_\theta h_t\left(z_{I_t}, \bar{\theta}\right)\\&= \left\langle\nabla_\theta^2 L, \Delta \theta\right\rangle+\left\langle\nabla_{\theta \lambda} L, \gamma\right\rangle+\left\langle\nabla_{\theta \eta_N} L, \Delta \eta_N\right\rangle.
\end{align}
The optimality condition~(\ref{35}) requires that:
\begin{equation}\label{38}
    \nabla_{\Delta \theta} L_{\text {aux }}(\Delta \theta, \gamma)= \left\langle\nabla_\theta^2 L, \Delta \theta\right\rangle+\left\langle\nabla_{\theta \lambda} L, \gamma\right\rangle+\left\langle\nabla_{\theta \eta_N} L, \Delta \eta_N\right\rangle=0.
\end{equation}
with $\gamma_g^j \geq 0$ when $j \in I_0$ having $\bar{g}_j\left(\Delta \theta, z_{I_j}\right)=0$. All other $\gamma_g^j$ are zero. This comes from the complementarity condition in the KKT conditions of the auxiliary problem.

It is straightforward to see that the solution $\Delta\theta$ to~(\ref{requirements}) is the same as the solution to~(\ref{38}). Thus, the auxiliary problem shares the same solution set as the auxiliary variational inequality. \qed

\section{Machine Unlearning of SVM}\label{svm appendix}
This section analyzes the sensitivity of the SVM solution with the data weight $\eta_N$. As introduced in Section~\ref{sec:4.1} , the data weighted SVM model is:
\begin{equation}\label{weighted SVM}
\begin{aligned}
\min_{\theta=\{w,b,\xi\}} \quad &\frac{1}{2}\|{w}\|^2 + C\sum_{i=1}^{N-1} \xi_i+\eta_N \cdot C \max\left[0,1-y_N\left({w}^\top x_N + b\right)\right] \\
\textrm { s.t       } \quad &y_i\left({w}^\top x_i + b\right) \geq 1 - \xi_i, \\
&\xi_i \geq 0. \\
&i=1, \ldots, N-1.
\end{aligned}
\tag{Data Weighted SVM}
\end{equation}
We use the parameterized Softplus function to replace the hinge loss $\max \left[0,1-y_N\left({w}^{\top} x_N+b\right)\right]$. The parameterized Softplus function is defined as: \begin{equation}
    \operatorname{Softplus}_\beta(1-y_N\left({w}^{\top} x_N+b\right))=\frac{1}{\beta} \ln \left(1+e^{\beta \left(1-y_N\left({w}^{\top} x_N+b\right)\right)}\right),
    \end{equation}
where $\beta>0$ controls the sharpness of the approximation. As $\beta \rightarrow \infty$, the Softplus function converges to the hinge loss, providing a smooth and differentiable approximation while maintaining numerical stability. 

The Lagrangian function of the model (\ref{weighted SVM}) is
\begin{equation}
\begin{aligned}
L({\eta_N},\theta,\alpha,\mu)= & \frac{1}{2}\|{w}\|^2+C \sum_{i=1}^{N-1} \xi_i+ \frac{\eta_N C}{\beta} \ln \left(1+e^{\beta\left(1-y_N\left({w}^{\top} x_N+b\right)\right)}\right) \\
& +\sum_{i=1}^{N-1} \alpha_i\left[y_i\left({w}^{\top} \mathbf{x}_i+b\right)-1+\xi_i\right]-\sum_{i=1}^{N-1} \mu_i \xi_i .
\end{aligned}
\end{equation}
Here, $\alpha=[\alpha_i]$ and $\mu=[\mu_i],i=1,2,\dots, N-1$, are Lagrange multipliers corresponding to the first and second constraints, respectively. Based on the values of $\alpha_i$, the training points are categorized as margin support vectors $(0<$ $\alpha_i<C$, set $S$ ), error support vectors ( $\alpha_i=C$, set $E$ ), and reserve points ( $\alpha_i=0$, set $R$ ). The sets $R$ and $E$ can be further subdivided based on the relationship between $y_i\left({w}^{\top} x_i+b\right)$ and the margin boundary:
\begin{equation}
R=\left\{\begin{array}{ll}
R_0 & \text { if } y_i\left({w}^{\top} x_i+b\right)-1=0, \\
R_1 & \text { if } y_i\left({w}^{\top} x_i+b\right)-1<0 .
\end{array} \quad E= \begin{cases}E_0 & \text { if } y_i\left({w}^{\top} x_i+b\right)-1=0, \\
E_1 & \text { if } y_i\left({w}^{\top} x_i+b\right)-1<0 .\end{cases}\right.
\end{equation}

When $\eta_N=\bar{\eta}_N=1$, we denote the solution as $\bar{\theta}=\{\bar{w}, \bar{b}, \bar{\xi}\}$ and  $\bar{\alpha}, \bar{\mu}$ as the optimal solution of the model (\ref{weighted SVM}). We assume the $\eta_N$ moves along the direction of $q$. The auxiliary problem is given by:
\begin{equation}\label{Aux SVM}
\begin{aligned}
\min_{\Delta \theta = (\Delta {w}, \Delta b, \Delta \boldsymbol{\xi})} \quad 
& \bar{g}_0(\Delta \theta) - \langle {\gamma}, \Delta \theta \rangle \\
\text{s.t.    } \quad 
& g_i(\Delta \theta):=1 - \xi_i - y_i \big({w}^\top x_i + b \big) -y_i\big( \Delta {w}^\top x_i+\Delta b \big) -\Delta \xi_i  
\begin{cases}
=0 & \text{if } i \in S \cup E_0 \cup E_1, \\
\leq 0 & \text{if } i \in R_0, \\
\text{free} & \text{if } i \in R_1,
\end{cases} \\
&h_i(\Delta \theta):= -\xi_i - \Delta \xi_i 
\begin{cases}
=0 & \text{if } i \in S \cup R_0 \cup R_1, \\
\leq 0 & \text{if } i \in E_0, \\
\text{free} & \text{if } i \in E_1.
\end{cases}
\\&i=1, 2, \dots, N-1
\end{aligned}
\tag{Auxiliary SVM}
\end{equation}
Where $\bar{g}_0(\Delta \theta)$ is the second-orden expansion of $L(\bar{\eta}_N, \bar{\theta}, \bar{\alpha}, \bar{\mu})$:
\begin{equation*}
\begin{aligned}
\bar{g}_0(\Delta \theta) 
&= L(\bar{\eta}_N, \bar{\theta}, \bar{\alpha}, \bar{\mu}) 
+ \nabla_\theta L(\bar{\eta}_N, \bar{\theta}, \bar{\alpha}, \bar{\mu}) \cdot \Delta \theta 
+ \frac{1}{2} \Delta \theta^\top \nabla_\theta^2 L(\bar{\eta}_N, \bar{\theta}, \bar{\alpha}, \bar{\mu}) \Delta \theta \\
&= L(\bar{\eta}_N, \bar{\theta}, \bar{\alpha}, \bar{\mu}) 
+ \frac{1}{2} \Delta {w}^\top \Delta {w} 
+ \frac{\bar{\eta}_N C \beta}{2}M\cdot(\Delta {w}^\top x_N)^2
+ C\bar{\eta}_N\beta \Delta b M\cdot(\Delta {w}^\top x_N)
+ C\bar{\eta}_N\beta M(\Delta b)^2,\\
\end{aligned}
\end{equation*}
\begin{equation}
\gamma=\left(\begin{array}{c}
q C y_N \sigma \cdot x_N \\\
q C \sigma  y_N \\
\mathbf{0} \in \mathbb{R}^{N-1}
\end{array}\right),
\end{equation}
\begin{equation}
\sigma= \frac{\exp\big(\beta (1 - y_N (\bar{w}^\top x_N + \bar{b})\big)}{1+\ln\big(1+ \exp\big(\beta (1 - y_N (\bar{w}^\top x_N + \bar{b})\big)\big)},
\end{equation}

\begin{equation}
{M}= \sigma(1-\sigma)
\end{equation}

We denote the Lagrangian multipliers of the auxiliary problem (\ref{Aux SVM}) for $g_i, h_i$ as $z^g_i,z^h_i$, respectively. Following (\ref{VI}), the first-order optimality condition of the auxiliary problem (\ref{Aux SVM}) w.r.t. $\theta, z_i^g$, and $z_i^h$ can be formulated as:
\begin{equation}\label{VI:SVM}
\begin{aligned}
&\begin{bmatrix}
(I + w C\beta{M}\cdot x_N x_N^\top) \Delta w + w C\beta\Delta b M \cdot{x_N}
- \sum_{i=1}^{N-1} z_i^g y_i x_i \\
w C\beta {M}\cdot \Delta w^\top x_N + w C\beta {M}\Delta b 
- \sum_{i=1}^{N-1} z_i^g y_i \\
-z_i^g - z_i^h \quad (i=1, 2, \ldots, N-1)
\end{bmatrix} 
- {v} = 0,
\end{aligned}
\end{equation}
The change of ${w}$, denoted by $\Delta{w}$, can be obtained by rearranging (\ref{VI:SVM}): 
\begin{equation}
\begin{aligned}
&\Delta {w}=\sum_{i=1}^{N-1} z_i^gy_i\left(\mathbf{x}_i-\mathbf{x}_N\right)
\end{aligned}
\end{equation}
The values of $z_i^g$ can be derived by solving the auxiliary problem (\ref{Aux SVM}). Using the complementary slackness condition of the auxiliary model, we can provide the following value ranges:
\begin{equation}
z_i^g= \begin{cases}0, & \text { if } i \in R_1 \cup E_1 \\ \geq 0, & \text { for } i \in R_0 \text { such that }-y_i\left(\mathbf{x}_i^{\top} \Delta \mathbf{w}+\Delta b\right)-\Delta \xi_i=0 \\ 0, & \text { for } i \in R_0 \text { such that }-y_i\left(\mathbf{x}_i^{\top} \Delta \mathbf{w}+\Delta b\right)-\Delta \xi_i<0 \\ free & \text { for } i \in S \cup E_0\end{cases}
\end{equation}

After identifying $\Delta w$, the value of $\Delta b$  can be derived from the following equation: 
\begin{equation}
\begin{aligned}
&y_i\left(\mathbf{x}_i^{\top} \Delta \mathbf{w}\right)+y_i \Delta b+\Delta \xi_i=0, \quad \text { for any } i \in S
\end{aligned}
\end{equation}

By solving the auxiliary problem (\ref{Aux SVM}), we can derive $\Delta \theta = \{\Delta w, \Delta b, \Delta \xi\}$ without retraining the SVM. 

\section{Multiple Data Points Deletion Scenario with Multi-Constraint Involvement}\label{Multi-Constraint appendix}
We consider the removal of a set of data points  \( {K} = \{z_{i_1}, z_{i_2}, \dots, z_{i_m}\} \). The corresponding gold standard solution \( \bar{\theta}_{-K} \) is obtained by retraining the ML-TSEP model~(\ref{orginal problem}) after excluding \( K \) from both the objective and the constraints. Let ${J}_{R}$ (resp. ${T}_{{R}}$ ) denote the indices of inequality (resp. equality) constraints involving any $z_i \in K$. The corresponding gold standard ML-TSEP is formulated as:
\begin{equation}\label{eq:multi_gold_standard}
\begin{aligned}
\bar{\theta}_{-{K}} = \arg \min_{\theta} & \sum_{z_i \notin K} \ell(z_i, \theta) \\
\text{s.t.} \quad 
& g_j(z_{I_j} \setminus K, \theta) \leq 0, \quad \forall j \in {J}_{{R}}, \\
& g_j(z_{I_j}, \theta) \leq 0, \quad \forall j \in \{1, \ldots, J\} \setminus {J}_{{R}}, \\
& h_t(z_{I_t} \setminus K, \theta) = 0, \quad \forall t \in {T}_{{R}}, \\
& h_t(z_{I_t}, \theta) = 0, \quad \forall t \in \{1, \ldots, T\} \setminus {T}_{{R}}.
\end{aligned}
\end{equation}

To approximate the gold standard solution \( \bar{\theta}_{-{K}} \) without retraining, we adopt a data-weighted formulation by introducing a weight \( \eta_i \in [0,1] \) for each \( z_i \in K \). Define the vector \( {\eta}_K = [\eta_{i_1}, \eta_{i_2}, \dots, \eta_{i_m}] \). When all weights are 1, the original ML-TSEP model is recovered; when all are 0, the points in \( K \) are removed. Following the model~(\ref{weighted penalty problem}), we have a data weighted TSEP formulated as:
\begin{equation}\label{eq:multi_weighted_problem}
\begin{aligned}
{\theta}({\eta}_K) = \arg \min_{\theta} & \sum_{z_i \notin K} \ell(z_i, \theta) + \sum_{z_i \in K} \eta_i \cdot \ell(z_i, \theta) \\
& + \sum_{j \in J_R} \phi\left(C_g, g_j(z_{I_j} \setminus K, {\eta}_K \cdot K, \theta)\right) \\
& + \sum_{t \in T_R} \{\phi\left(C_h, h_t(z_{I_t} \setminus K, {\eta}_K \cdot K, \theta)\right) + \phi\left(C_h, -h_t(z_{I_t} \setminus K, {\eta}_K\cdot K, \theta)\right)\} \\
\text{s.t.}\quad 
& g_j(z_{I_j}, \theta) \leq 0, \quad \forall j \in \{1, \ldots, J\} \setminus J_R, \\
& h_t(z_{I_t}, \theta) = 0, \quad \forall t \in \{1, \ldots, T\} \setminus T_R.
\end{aligned}
\end{equation}

Let \( L( {\eta}_K,\theta, \lambda) \) denote the Lagrangian:
\begin{equation}\label{eq:multi_lagrangian}
\begin{aligned}
L( {\eta}_K,\theta,  \lambda) &= \sum_{z_i \notin K} \ell(z_i, \theta) + \sum_{z_i \in K} \eta_i \cdot \ell(z_i, \theta) \\
&+ \sum_{j \in J_R} \phi\left(C_g, g_j(z_{I_j} \setminus K, {\eta}_K\cdot K, \theta)\right) + \sum_{t \in T_R} \phi\left(C_h, h_t(z_{I_t} \setminus K, {\eta}_K\cdot K, \theta)\right) \\
&+ \sum_{t \in T_R} \phi\left(C_h, -h_t(z_{I_t} \setminus K, {\eta}_K\cdot K, \theta)\right) \\
&+ \sum_{j \in \{1,\ldots,J\} \setminus J_R} \lambda_g^j g_j(z_{I_j}, \theta) + \sum_{t \in \{1,\ldots,T\} \setminus T_R} \lambda_h^t h_t(z_{I_t}, \theta).
\end{aligned}
\end{equation}

We define \( \bar{\theta} = {\theta}({{\eta}}_K = \boldsymbol{1}) \), \( \bar{\lambda} \) as its Lagrange multiplier, and define \( q = -\boldsymbol{1} \in \mathbb{R}^m \) as the perturbation direction for $\eta_K$ from $\bar{\eta}_k=1$. Then the auxiliary problem solves for \( \Delta\theta \):

\begin{equation}\label{eq:multi_aux_problem}
\begin{aligned}
\min_{\Delta\theta} \quad 
& \bar{g}_0(\Delta\theta) + \left\langle \nabla_{\theta {\eta}_k}^2 L( \bar{{\eta}}_K, \bar{\theta},  \bar{\lambda}) \cdot \boldsymbol{q}, \Delta\theta \right\rangle \\
\text{s.t.} \quad 
& \bar{g}_j(\Delta\theta, z_{I_j}) = 0, \quad \forall j \in \tilde{I} \setminus \tilde{I}_0 \\
& \bar{g}_j(\Delta\theta, z_{I_j}) \leq 0, \quad \forall j \in \tilde{I}_0 \\
& \bar{g}_j(\Delta\theta, z_{I_j}) \text{ unrestricted}, \quad \forall j \in \tilde{I}_1 \\
& \bar{h}_t(\Delta\theta, z_{I_t}) = 0, \quad \forall t \in \{1, \ldots, T\}\setminus T_R
\end{aligned}
\end{equation}

Here, \( \bar{g}_0(\Delta\theta) \) is the second-order expansion of the Lagrangian function:
\begin{equation}
\bar{g}_0(\Delta\theta) = L(\bar{{\eta}}_K, \bar{\theta}, \bar{\lambda}) + \left\langle \nabla_\theta L(\bar{{\eta}}_K, \bar{\theta}, \bar{\lambda}), \Delta\theta \right\rangle + \frac{1}{2} \Delta\theta^\top \nabla_{\theta\theta}^2 L(\bar{{\eta}}_K, \bar{\theta}, \bar{\lambda}) \Delta\theta
\end{equation}

The linearized constraints are:
\begin{equation}
\begin{aligned}
\bar{g}_j(\Delta\theta, z_{I_j}) &= g_j(z_{I_j}, \bar{\theta}) + \left\langle \nabla_\theta g_j(z_{I_j}, \bar{\theta}), \Delta\theta \right\rangle \\
\bar{h}_t(\Delta\theta, z_{I_t}) &= h_t(z_{I_t}, \bar{\theta}) + \left\langle \nabla_\theta h_t(z_{I_t}, \bar{\theta}), \Delta\theta \right\rangle
\end{aligned}
\end{equation}

We define the index sets as follows:
\begin{equation}
\begin{aligned}
\tilde{I} &= \left\{ j \in \{1, \ldots, J\}\setminus J_R \mid g_j(\bar{\theta}, z_{I_j}) = 0 \right\} \\
\tilde{I}_0 &= \left\{ j \{1, \ldots, J\}\setminus J_R  \mid g_j(\bar{\theta}, z_{I_j}) = 0,\bar{\lambda}_g^j = 0 \right\}\subset \tilde{I} \\
\tilde{I}_1 &= \left\{ j \in \{1, \ldots, J\}\setminus J_R  \mid g_j(\bar{\theta}, z_{I_j}) < 0 \right\}
\end{aligned}
\end{equation}

By solving the auxiliary problem~\eqref{eq:multi_aux_problem}, which is also a quadratic program, we obtain a local linear approximation to the solution change when removing multiple data points \( {K} \).

\newpage
\bibliographystyle{apalike} 
\bibliography{reference}
\end{document}